%% file: main.tex
\documentclass{article}
\input{preamble}

\title{SearchGym: Bootstrapping Real-World Search Agents via Cost-Effective and High-Fidelity Environment Simulation}

\author[1]{Xichen Zhang}
\author[2]{Ziyi He}
\author[2]{Yinghao Zhu}
\author[3]{Sitong Wu}
\author[3]{Shaozuo Yu}
\author[1]{Meng Chu}
\author[1]{Wenhu Zhang}
\author[2]{Haoru Tan}
\author[1,$\dagger$]{Jiaya Jia}

\affil[1]{The Hong Kong University of Science and Technology}
\affil[2]{The University of Hong Kong}
\affil[3]{The Chinese University of Hong Kong}

\abstract{%
Search agents have emerged as a pivotal paradigm for solving open-ended, knowledge-intensive reasoning tasks. However, training these agents via Reinforcement Learning (RL) faces a critical dilemma: interacting with live commercial Web APIs is prohibitively expensive, while relying on static data snapshots often introduces noise due to data misalignment. This misalignment generates \textit{corrupted reward signals} that destabilize training by penalizing correct reasoning or rewarding hallucination. To address this, we propose \textbf{SearchGym}, a simulation environment designed to bootstrap robust search agents. SearchGym employs a rigorous generative pipeline to construct a verifiable knowledge graph and an aligned document corpus, ensuring that every reasoning task is factually grounded and strictly solvable. Building on this controllable environment, we introduce \textbf{SearchGym-RL}, a curriculum learning methodology that progressively optimizes agent policies through purified feedback, evolving from basic interactions to complex, long-horizon planning. Extensive experiments across the Llama and Qwen families demonstrate strong Sim-to-Real generalization. Notably, our Qwen2.5-7B-Base model trained within SearchGym surpasses the web-enhanced ASearcher baseline across nine diverse benchmarks by an average relative margin of 10.6\%. Our results validate that high-fidelity simulation serves as a scalable and highly cost-effective methodology for developing capable search agents.
}

\footnotemarkers{$^\dagger$Corresponding author. \faEnvelope\ Xichen Zhang: \texttt{xichenzhang879@gmail.com}}

\codelink{https://github.com/JIA-Lab-research/SearchGym}


\begin{document}
\maketitle

\section{Introduction}
Autonomous search agents, empowered by Large Language Models (LLMs) to interact with external search engines, represent a pivotal advancement in solving complex, knowledge-intensive tasks~\cite{yao2022react,jin2025search,sun2025zerosearch,wang2024survey}. Unlike passive retrieval systems, these agents can autonomously formulate queries, browse results, and synthesize information over multiple turns~\cite{gao2025beyond}. To master such sophisticated decision-making processes, Reinforcement Learning (RL) has emerged as the standard training paradigm, enabling models to optimize their reasoning trajectories through outcome-based feedback~\cite{DeepSeekR1,openaio1}.

However, the advancement of search agents is fundamentally constrained by the training environment. Researchers currently face a critical dilemma between scalability and realism. On one hand, training directly with live commercial Web APIs offers high environmental fidelity but incurs prohibitive costs. As demonstrated in Table~\ref{tab:main_results_7b_challenging}, conducting large-scale RL experiments is financially unsustainable; a single training epoch over 30,000 questions, with 8 rollouts per question and an average of 3 search actions per rollout, necessitates approximately 720,000 API calls. At standard commercial rates, this accumulates fees exceeding \$500 per run, rendering extensive exploration and hyperparameter tuning impractical. On the other hand, relying on static offline assets such as Wikipedia snapshots introduces significant noise due to inherent data misalignment. This noise manifests as outdated information, ambiguous queries, and factual inconsistencies between the ground truth and the retrieval corpus. Our analysis identifies these issues as the source of corrupted reward signals. Consequently, the optimization process is destabilized because agents are frequently penalized for correct reasoning or rewarded for hallucinations that coincidentally match the labels, leading to the training instability and policy collapse (Figure~\ref{fig:Compa}).

To bridge this gap, we introduce \textbf{SearchGym}, a cost-effective and high-fidelity simulation environment designed to bootstrap robust search agents without the costs of live web access. To resolve the training instability caused by noisy training data, SearchGym constructs a verified knowledge graph and an aligned document corpus in a generative closed-loop environment. This guarantees that every reasoning task is factually grounded and strictly solvable. Within this controlled environment, we propose \textbf{SearchGym-RL}, a training methodology incorporating curriculum learning to optimize agent policies. By leveraging clean, noise-free signals, SearchGym-RL progressively guides agents from mastering basic interactions to handling complex reasoning structures. Furthermore, by training on synthetic data unseen during pretraining, we compel the agent to rely exclusively on tool execution rather than parametric memory, fostering robust search capabilities that generalize to the real world. Our main contributions are:
\begin{enumerate}
    \item \textit{Insight:} We identify \textit{corrupted reward signals} stemming from data misalignment in existing offline training, demonstrating that high-fidelity reward signals are a necessary condition for stable RL-based search agent training.
    \item \textit{Methodology:} We introduce SearchGym, a closed-loop environment constructed via a rigorous pipeline spanning from knowledge graph synthesis to solvable QA generation. Leveraging this environment, we employ SearchGym-RL, a curriculum-based strategy that guarantees task solvability and provides purified feedback, enabling agents to master complex reasoning structures (e.g., compositional and parallel search) efficiently.
    \item \textit{Experiment:} We demonstrate robust Sim-to-Real generalization across diverse model families, including Llama 3.2, Qwen 2.5, and Qwen 3. Despite being trained in a synthetic environment with zero commercial web API costs, our Qwen2.5-7B agent surpasses the web-enhanced ASearcher baseline across 9 diverse benchmarks by an average relative margin of 10.6\%. Our comprehensive experiments encompass both standard QA tasks and challenging, open-ended research benchmarks such as GAIA and xbench-DeepSearch, validating the efficacy of SearchGym and SearchGym-RL on complex real-world problems.
\end{enumerate}

\section{Related Work}

\paragraph{Search agents and tool learning.}
The paradigm of augmenting Large Language Models (LLMs) with external knowledge has shifted from passive Retrieval-Augmented Generation (RAG), where the model receives a fixed set of retrieved context, to iterative, agent-based workflows~\cite{gao2023retrieval,lewis2020retrieval}. Early approaches rely on prompt engineering frameworks, such as ReAct~\cite{yao2022react}, or Supervised Fine-Tuning (SFT)~\cite{schick2023toolformer} to teach models how to utilize search tools. While SFT effectively bootstraps basic tool-use capabilities, it remains limited by the quality of human-annotated trajectories. Consequently, Reinforcement Learning (RL)~\cite{DeepSeekR1,jin2025search,sun2025zerosearch,gao2025beyond} has emerged as a standard for optimizing complex interaction policies, enabling agents to autonomously discover search strategies through outcome-based feedback. However, while frameworks such as ASearcher~\cite{gao2025beyond} validate the efficacy of training against live web engines, their reliance on commercial APIs incurs prohibitive financial costs, rendering large-scale policy optimization unsustainable.

\paragraph{Simulated search environments.}
To mitigate the prohibitive costs of online training, recent research has pivoted toward offline or simulated environments. Approaches such as Search-R1~\cite{jin2025search} establish a local retrieval environment based on the 2018 Wikipedia corpus to facilitate training on datasets NQ~\cite{kwiatkowski2019natural} and HotpotQA~\cite{yang2018hotpotqa}. However, this paradigm faces two limitations: the static corpus largely overlaps with the pretraining data of modern LLMs, and the data quality within these benchmarks varies significantly. Alternatively, ZeroSearch~\cite{sun2025zerosearch} employs an LLM to simulate search engine responses. While innovative, this approach is often insufficient for training models on complex, long-horizon reasoning tasks. In contrast, SearchGym generates a grounded, verifiable, and closed-loop synthetic world. By constructing tasks from a validated knowledge graph, SearchGym ensures high-fidelity interactions and guarantees solvability for multi-hop reasoning.

\begin{figure*}[!ht]
\centering
\includegraphics[width=1\linewidth]{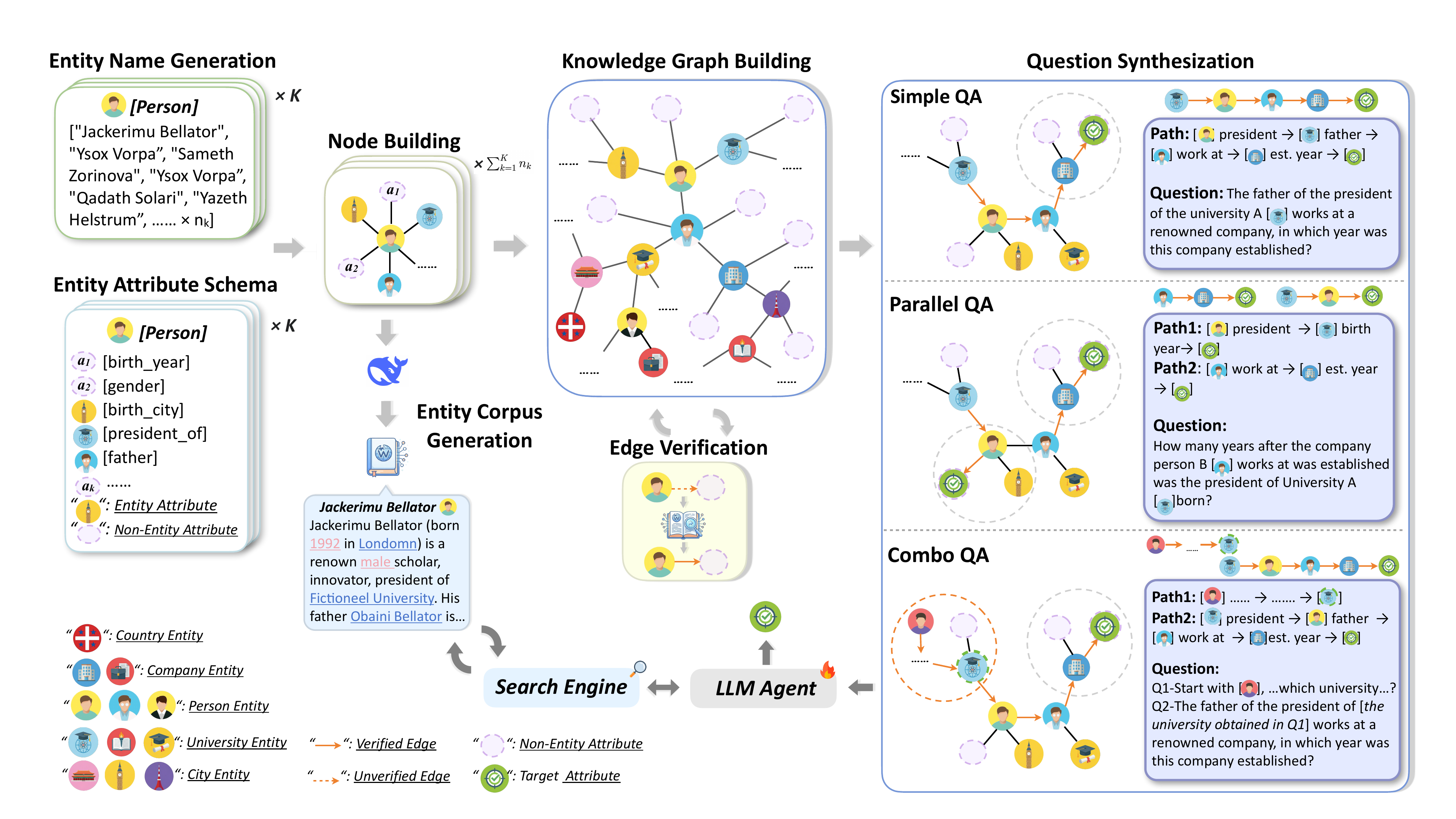}
\caption{An overview of the SearchGym pipeline. From a verified knowledge graph (left), we sample reasoning paths of varying structures. These paths are verbalized into complex, solvable Question-Answering pairs (right), categorized as Simple, Parallel, or Combo QA. This creates a closed-loop environment where the LLM agent interacts with a search engine to retrieve evidence and solve the tasks. The legend defines the visual elements.}
\label{fig:data_pipeline}
\end{figure*}

\section{Methodology}

SearchGym constitutes a high-fidelity and cost-effective offline simulation designed for training search agents. As illustrated in Figure~\ref{fig:data_pipeline}, our four-stage pipeline constructs a verifiable knowledge graph and aligned corpus, ensuring every task is logically consistent and strictly solvable. Leveraging this controlled environment, we propose {SearchGym-RL}, a curriculum-based reinforcement learning framework that progressively optimizes agent policies through purified feedback.

\subsection{SearchGym Simulation Environment}

To address the lack of rigor in existing offline methods, we formalize the data generation process. Let the world model be defined as a tuple $\mathcal{W} = \langle \mathcal{G}, \mathcal{D} \rangle$, where $\mathcal{G}$ is a structured knowledge graph and $\mathcal{D}$ is the associated searchable corpus. All LLM prompts employed in our data generation process and generation details are provided in Appendix~\ref{sec:Data Generation Details}.

\paragraph{Knowledge graph synthesis.}
First, we construct a knowledge graph $\mathcal{G} = (\mathcal{V}, \mathcal{E})$ based on a predefined schema $\mathcal{S}$. The schema defines valid entity types (e.g., {Person}, {Organization}) and relational constraints. We programmatically populate the vertex set $\mathcal{V}$ with diverse, fictional yet plausible entities and establish directed edges $\mathcal{E}$ to represent semantic relationships. This construction ensures that $\mathcal{G}$ exhibits a realistic topological structure while remaining fully controllable.

\paragraph{Entity corpus generation.}
We construct a searchable document corpus $\mathcal{D} = \{d_v \mid v \in \mathcal{V}\}$. For each entity $v$, we employ an LLM $M_{\text{gen}}$ to synthesize a Wikipedia-style document $d_v = M_{\text{gen}}(v, \mathcal{N}_v)$, which incorporates the entity's attributes and its local graph neighborhood $\mathcal{N}_v$. To simulate a web environment, each document $d_v$ is associated with a unique URL, allowing agents to perform distinct {Search} and {Access} actions.

\paragraph{Edge verifiability and filtering.}
A critical innovation of SearchGym is the enforcement of learnability through retrievability. For every edge $e=(u, v) \in \mathcal{E}$, we generate search queries $\mathcal{Q}_e$ (specifically, we set $|\mathcal{Q}_e|=15$). We execute these queries against $\mathcal{D}$ using a retrieval engine $\mathcal{R}$. An edge is retained in the verified subgraph $\mathcal{G}^* \subseteq \mathcal{G}$ if and only if the target document $d_v$ is retrieved by at least 5 of these queries:

{
\begin{equation}
    (u, v) \in \mathcal{E}^* \iff \left| \{ q \in \mathcal{Q}_e \mid d_v \in \text{Top-}K(\mathcal{R}(q, \mathcal{D})) \} \right| \geq 5.
\end{equation}
}
This filtering ensures the reasoning path is discoverable via search, decoupling the agent's reasoning capability from stochastic retrieval failures and ensuring clean training signals.

\paragraph{Question-answer synthesis from verified paths.}
Finally, we synthesize complex Question-Answering (QA) pairs by sampling reasoning paths from $\mathcal{G}^*$. A path $\mathcal{P} = (v_0, e_1, v_1, \dots, v_k)$ represents a logic chain of length $k$. We categorize paths into three structural types:
\begin{enumerate}
    \item \textbf{Simple QA:} Linear paths where $v_i \to v_{i+1}$.
    \item \textbf{Parallel QA:} Multiple independent paths $\mathcal{P}_1, \mathcal{P}_2$ merging at a terminal condition, requiring information synthesis.
    \item \textbf{Combo QA:} Nested paths where the answer to $\mathcal{P}_1$ parametrizes the query for $\mathcal{P}_2$.
\end{enumerate}
The generator $M_{\text{gen}}$ verbalizes these paths into natural language questions $Q$ and answers $A$, guaranteeing every task $(Q, A)$ is solvable within $\mathcal{D}$.

\paragraph{Data statistics.}
The synthesized environment consists of approximately 3,600 nodes, each corresponding to a generated document. From this corpus, our pipeline yields over 41,000 unique QA pairs, stratified by complexity as shown in Table~\ref{tab:data_distribution}. For evaluation, we sample a representative subset of 642 questions (462 Simple, 134 Parallel, 46 Combo). We designate this high-quality evaluation set as {SearchGymBench}. These instances span 1 to 12 hops, explicitly testing the capacity for sustained, long-horizon reasoning.

\input{tables/data_statistics}

\subsection{SearchGym-RL Training}

Our methodology focuses on training search agents within the SearchGym environment to master a realistic action space consisting of {Search} and {Access}. This process is driven by two core components: a reinforcement learning framework for policy optimization, and a curriculum learning strategy designed to progressively enhance agent capabilities.

\paragraph{Action Space.} 
To mimic web browsing and decouple retrieval from reading, we define two primitives: \textbf{Search($q$)} returns a list of snippets with URLs, while \textbf{Access($u$)} retrieves the full document content $d_u$. This design not only simulates web browsing but also enforces rigorous reasoning, compelling the agent to evaluate relevance based on summaries before committing to detailed reading.

\paragraph{Reinforcement learning within SearchGym.}

Our agent addresses problems via a multi-turn reasoning process, generating a trajectory $\mathcal{T}$ of intermediate thoughts and actions that culminates in a final answer. Unlike single-step generation, this iterative process allows for complex problem decomposition. 
The quality of each trajectory $\mathcal{T}$ is measured by a terminal reward $R(\mathcal{T})$, determined by the correctness of the final answer. Specifically, we define the reward using the token-level F1 score, which assesses the token overlap between the prediction and the ground truth. 

To optimize the policy $\pi_\theta$, we employ Group Relative Policy Optimization (GRPO)~\cite{shao2024deepseekmath}, building on recent advancements~\cite{jin2025search,sun2025zerosearch,gao2025beyond}. GRPO operates on a group of $N$ trajectories $\{\mathcal{T}_i\}_{i=1}^N$ sampled from the policy. For each trajectory, it computes a normalized advantage $\hat{A}_i$ by standardizing the terminal F1 rewards within the group. 
The policy parameters $\theta$ are then updated by maximizing the following clipped surrogate objective:
\begin{equation} \label{eq:grpo_simple}
\mathcal{L}(\theta) = \mathbb{E}_{\tiny \mathcal{T}_i \sim \pi_{\theta_{\text{old}}}} \Big[ \min\left(\rho_i(\theta)\hat{A}_i, \text{clip}(\rho_i(\theta), 1-\epsilon, 1+\epsilon)\hat{A}_i\right) - \beta D_{KL}(\pi_\theta(\mathcal{T}_i) \| \pi_{\text{ref}}(\mathcal{T}_i)) \Big]
\end{equation}
where $\rho_i(\theta) = \pi_\theta(\mathcal{T}_i) / \pi_{\theta_{\text{old}}}(\mathcal{T}_i)$ denotes the importance sampling ratio, $\epsilon$ is the clipping hyperparameter, and the $D_{KL}$ term imposes a penalty against a reference policy $\pi_{\text{ref}}$ to stabilize training.

\paragraph{Hierarchical skill acquisition via curriculum} 
Training agents on complex, multi-hop tasks from scratch is challenging due to sparse rewards, which can destabilize the learning process. To address this, we employ a two-stage curriculum that leverages the structured difficulty of tasks within SearchGym to progressively build agent capabilities.

\begin{itemize}
    \item \textit{Stage 1: Foundational Skill Acquisition.}
    We first train the agent on Simple QA tasks with shorter reasoning chains (1-6 hops). This stage focuses on mastering core skills, such as query formulation, document parsing, and sequential evidence, ensuring a stable start to training.
    \item \textit{Stage 2: Advanced Reasoning Development.}
    Once the agent reaches a baseline proficiency, we introduce more complex Parallel and Combo QA tasks and increase the proportion of long-horizon problems (6-12 hops). This challenges the agent to develop advanced abilities like problem decomposition and information synthesis, preparing it for real-world complexity.
\end{itemize}

\section{Experimental Setups}

\paragraph{Training datasets.}
We optimize the policy using the SearchGym Synthetic Corpus, comprising approximately 41,000 verifiable tasks. We strictly isolate the held-out {SearchGymBench} from training to ensure that performance reflects generalized reasoning capabilities rather than memorization.

\paragraph{Search tools.}
Following ASearcher~\cite{gao2025beyond}, we employ three distinct retrieval settings.
First, the {SearchGym Environment} utilizes a local index of our synthetic corpus $\mathcal{D}$ for training process.
Second, the {Local Wikipedia} setting employs the 2018 snapshot~\cite{kwiatkowski2019natural} for standard QA evaluation.
Third, the {Live Web} setting integrates a commercial Search API to address open-ended, challenging benchmarks.
Detailed configurations are provided in Appendix~\ref{ssec:appendix_search_tools}.

\paragraph{Baselines.}
To evaluate the effectiveness of our approach, we compare it against the following baselines:
(1) {Inference without Retrieval}: Base instruct models performing direct generation without external tools.
(2) {Inference with Retrieval}: Standard retrieval-augmented generation (RAG)~\cite{lewis2020retrieval}.
(3) {RL-Based Methods}: Search-R1~\cite{jin2025search}, ZeroSearch~\cite{sun2025zerosearch} and ASearcher~\cite{gao2025beyond}. We conduct these comparisons across a diverse suite of backbones, including the Qwen 2.5 series (1.5B/7B, Base/Instruct), Qwen 3 (4B/8B), and Llama 3.2 3B Instruct, to demonstrate that our method's efficacy is robust to scale and transferable across architectures.
\paragraph{Evaluation benchmarks.}
Our evaluation spans 10 diverse benchmarks to comprehensively assess agent capabilities. We categorize these into:
(1) {Single-hop QA}: Natural Questions (NQ)~\cite{kwiatkowski2019natural}, TriviaQA~\cite{joshi2017triviaqa}, and PopQA~\cite{mallen2023not}, testing factual recall and simple lookup.
(2) {Multi-hop QA}: 2WikiMultiHopQA(2Wiki)~\cite{ho2020constructing}, HotpotQA~\cite{yang2018hotpotqa}, Bamboogle~\cite{press2023measuring}, and Musique~\cite{trivedi2022musique}, evaluating complex reasoning chains.
(3) {Deep Research}: GAIA~\cite{mialon2023gaia} and xbench-DeepSearch(xbench)~\cite{chen2025xbench}, which serve as proxies for open-ended, real-world research tasks. 
(4) {Pure Tool-Use}: {SearchGymBench}, our held-out test suite comprising 642 complex tasks that are unseen during pretraining, thereby strictly evaluating tool-use capabilities rather than parametric memory.
Following established protocols~\cite{gao2025beyond}, for Bamboogle, GAIA, and xbench-DeepSearch, we use their full test sets. For GAIA specifically, we utilize the 103 examples from the text-only validation subset~\cite{li2025search}. For all other benchmarks, we use 1,000 randomly sampled instances.

\paragraph{Evaluation metrics.}
Following ASearcher's evaluation protocol~\cite{gao2025beyond}, we employ their robust LLM-as-a-Judge to assess semantic correctness given the open-ended tasks. Specifically, Qwen-2.5-72B-Instruct serves as the universal evaluator to verify consistency with the ground truth. We report Pass@1 for standard QA and Pass@4 for complex, long-horizon tasks such as GAIA and xbench-DeepSearch. Detailed evaluation prompts and protocols are provided in Appendix~\ref{ssec:evaluation_metrics}.

\paragraph{Implementation details.}
We train all models for 5 epochs using the AReal framework~\citep{fu2025areal}, reporting results from the best-performing checkpoint. For both training and evaluation, the retrieval module returns the top-5 relevant documents for every search action. We set the maximum interaction turns to 16 for training and 64 for evaluation. For training, we employ a maximum token limit of 1,024 with temperature 1.0 during training, whereas evaluation uses 4,096 tokens, temperature 0.6, and Top-$p$ 0.95. Consistent with recent studies~\citep{gao2025beyond}, we keep all other hyperparameters unchanged. A comprehensive list of hyperparameters is detailed in Appendix~\ref{ssec:hyperparameters}. 
\input{tables/main_result}
\section{Experimental Results}
We present a comprehensive evaluation of SearchGym-RL across diverse benchmarks and model architectures. The results demonstrate that agents trained in our high-fidelity offline simulation not only master standard retrieval tasks but also generalize effectively to complex, open-ended research problems without requiring expensive online interactions during training.

\paragraph{Performance on standard QA benchmarks.}
Table~\ref{tab:main_results_styled} details the performance on single-hop and multi-hop QA datasets. Our method consistently outperforms all baseline approaches. Notably, we achieve a relative improvement of 10.4\% over Search-R1, which relies on static Wikipedia snapshots under the Qwen-2.5-7B-Base model. This improvement confirms that the clean and verifiable logic in SearchGym provides effective training signals compared to noisy static datasets.

\paragraph{Robustness across model architectures and scales.}
SearchGym-RL maintains a significant lead over baselines regardless of the underlying model family (Qwen, Llama) or size (3B to 8B). As shown in Table~\ref{tab:main_results_styled}, on the Llama-3.2-3B backbone, SearchGym-RL achieves an average score of 54.96, representing a relative improvement of approximately 80\% over both Search-R1 and ZeroSearch. This demonstrates that the high-fidelity data from SearchGym is a universally effective driver of performance.

\paragraph{Sim-to-real generalization.}
Table~\ref{tab:main_results_7b_challenging} confirms that policies optimized within SearchGym generalize robustly to real-world web environments. Our Qwen-2.5-7B agent outperforms the web-trained ASearcher baseline by absolute margins of 3.89\% on GAIA and 17.00\% on xbench-DeepSearch. Crucially, our agent exhibits better efficiency, resolving complex tasks with 37.3\% fewer search actions (3.71 vs. 5.92) per query. These results validate high-fidelity simulation as a better methodology for developing autonomous search capabilities, yielding agents that excel in both accuracy and cost.

\paragraph{Extreme cost-effective.}
Beyond better performance, SearchGym offers a paradigm shift in training economics. As detailed in Table~\ref{tab:main_results_7b_challenging}, while the web-based ASearcher baseline incurs commercial API costs exceeding \$500 per run, our method achieves better results with {zero} API cost. This eliminates the financial barrier to training large-scale search agents, proving that high-performance autonomous systems can be forged without reliance on expensive external services.

\input{tables/challenge_table}

\paragraph{Scalability and data efficiency.}
We investigate the scaling properties of SearchGym-RL specifically within the second curriculum stage. To ensure a fair comparison, all models in this phase are optimized for a fixed 400 steps, and we report results from the best-performing checkpoint.

\begin{figure}[!ht]
\centering
\includegraphics[width=0.8\linewidth]{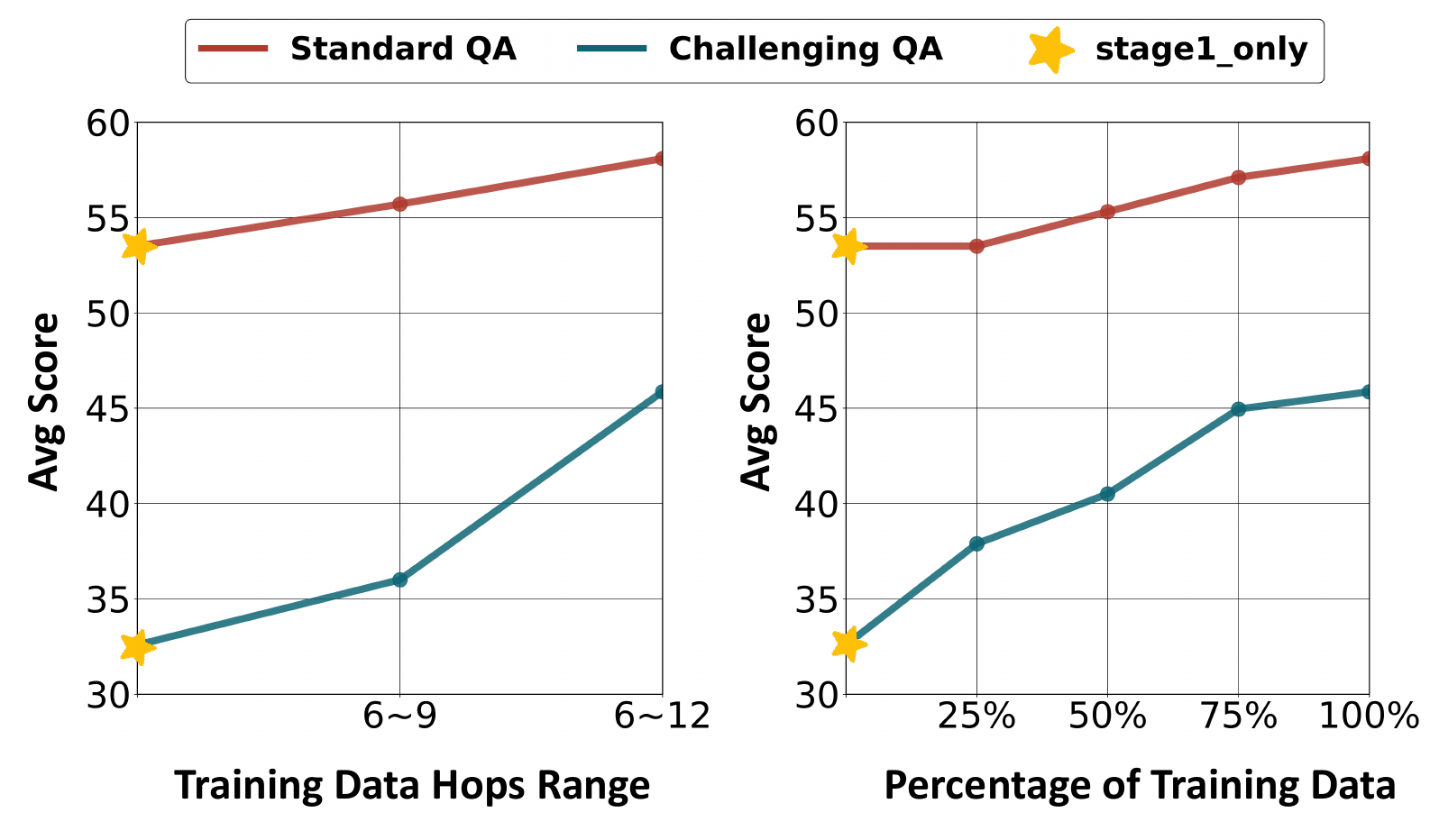}
\caption{Scalability analysis of SearchGym-RL. {Left:} Impact of maximum reasoning depth (hops) on downstream performance. {Right:} Performance trajectories across varying percentages of the training corpus. {Standard QA} represents the average score across single and multi-hop benchmarks; {Challenging QA} denotes the average on GAIA and xbench-DeepSearch.}
\label{fig:tendency}
\end{figure}

\noindent\textit{(1) Impact of reasoning complexity.}
Figure~\ref{fig:tendency} (Left) reveals that extending the maximum reasoning depth from 6 to 12 hops yields substantial gains on Challenging QA tasks (GAIA and Xbench), whereas moderate extensions (6--9 hops) provide limited benefit. This indicates that exposure to long-horizon synthetic logic acts as a prerequisite for unlocking the capability to solve open-ended real-world problems.

\noindent\textit{(2) Data efficiency and diversity.}
Simultaneously, the data coverage analysis in Figure~\ref{fig:tendency} (Right) demonstrates a strictly monotonic performance improvement as the corpus diversity increases from 0\% to 100\%, showing no signs of saturation. This positive trend validates SearchGym as a scalable data engine; simply scaling the volume of verified synthetic paths offers a deterministic pathway to forging increasingly capable agents.

\begin{figure*}[!ht]
\centering
\includegraphics[width=1\linewidth]{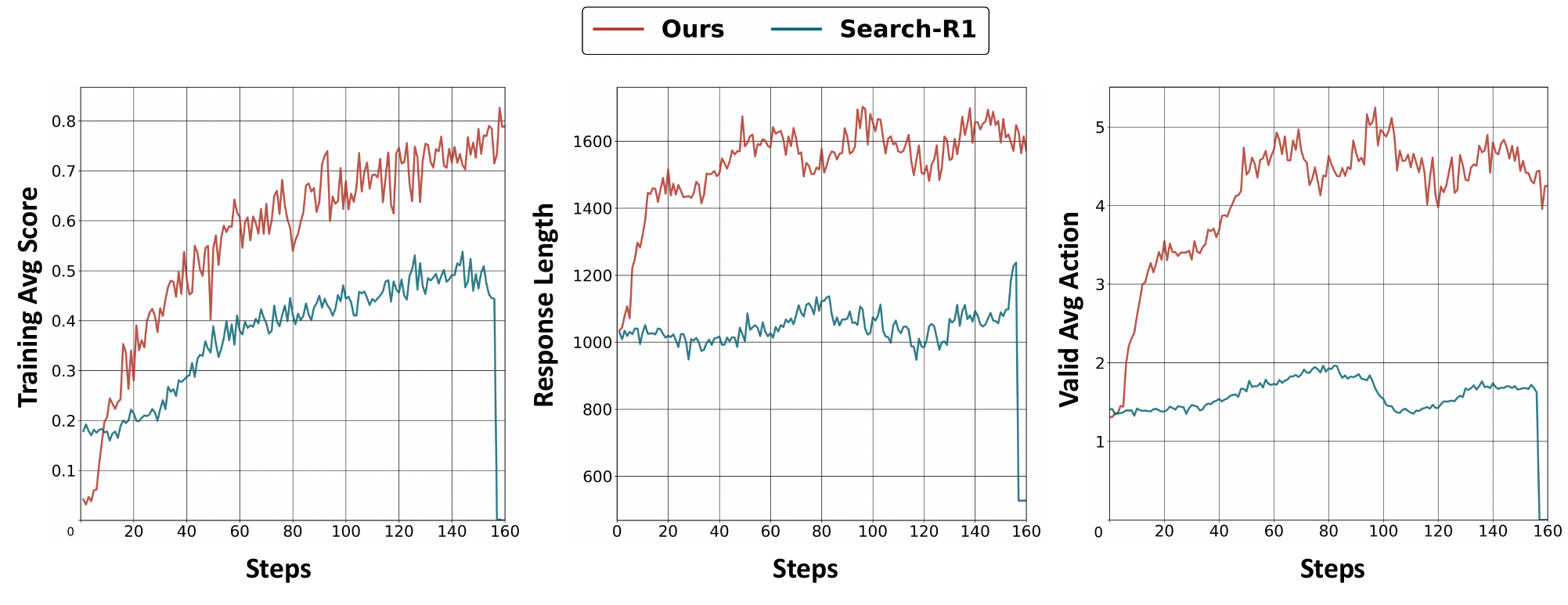}
\caption{Comparison of training dynamics between our method stage one and Search-R1. The Training Avg Score (left) represents the mean reward achieved for each rollout within a training batch. Our approach demonstrates stable, monotonic convergence towards a high reward, whereas Search-R1 exhibits significant volatility and eventual policy collapse. Note that the Search-R1 training curves are derived from their official public logs~\cite{searchr1_wandb}.}
\label{fig:Compa}
\end{figure*}

\input{tables/ablation_action}

\subsection{Ablation Study}
We perform an ablation study to isolate the contribution of key design components. Table~\ref{tab:ablation} summarizes the results using Qwen2.5-7B-Base.

\paragraph{Impact of action space on reasoning capabilities.}
We analyze how action space granularity influences agent behavior by comparing our dual-primitive framework ({Search} and {Access}) against a simplified {Search-only} baseline. In the baseline setting, the environment automatically returns the full content of retrieved documents immediately following a query, effectively removing the agent's agency in document selection. As shown in Table~\ref{tab:ablation}, this simplification leads to a significant performance decline, with a relative drop of approximately 19\% on the GAIA benchmark. We attribute this to the fact that a richer action space imposes a higher reasoning demand; the agent must actively evaluate the utility of search snippets and deliberately select which sources to investigate.

\paragraph{Importance of curriculum learning.}
We investigate the contribution of the two-stage training strategy by comparing our full method against two ablations: {Foundational Only} (omitting Stage 2) and {Mixed Training} (combining all tasks without staging). As shown in Table~\ref{tab:ablation}, removing Stage 2 results in a precipitate performance decline on challenging benchmarks, with the GAIA score dropping significantly from 42.72\% to 28.16\%. This evidence indicates that exposure to synthetic compositional structures, such as Parallel and Combo QA, serves as a strict prerequisite for generalizing to open-ended, long-horizon planning tasks. Furthermore, the explicit curriculum outperforms the Mixed Training baseline (52.09\% vs. 47.04\% on average). This confirms that a progressive difficulty gradient is essential for stable policy optimization.

\input{tables/benchmark}

\subsection{Further Analysis}

\paragraph{Performance on SearchGymBench.}
As shown in Table~\ref{tab:main_results}, this unseen synthetic dataset isolates tool-use capabilities from parametric memory. DeepSeek-V3.2 leads with 47.6\% accuracy on Complex QA, while baselines like ASearcher-Web-QwQ struggle (17.8\%). Our Qwen3-8B reaches 40.6\%, confirming the effective acquisition of robust reasoning primitives.

\paragraph{Training stability and signal purity.}
A critical advantage of SearchGym over methods utilizing static snapshots is the elimination of false negative rewards. Figure~\ref{fig:Compa} contrasts the training dynamics of our approach against Search-R1. We observe that Search-R1 suffers from significant volatility and eventual policy collapse after approximately 160 steps. We attribute this instability to the inherent data quality issues which lead to valid reasoning traces being penalized (corrupted reward signals), as detailed in Appendix~\ref{sec:search-r1_data_quality}. In contrast, SearchGym guarantees that every generated question can provide a correct reward signal within the environment. Consequently, our method exhibits a stable, monotonic improvement in reward, allowing the policy to converge to a higher performance.

\paragraph{Qualitative case study.}
We visualize comparative trajectories in Figure~\ref{fig:case_study} and Figure~\ref{fig:case_study_2}. Our agent demonstrates robust planning capabilities and effectively follows complex reasoning chains. In contrast, baselines frequently fail to extract the correct information from search results or resort to hallucination. Detailed analyses are in Appendix~\ref{sec:case_studies_appendix}.

\section{Conclusion}
We introduce SearchGym, a high-fidelity simulation that resolves the trade-off between offline scalability and online realism. By eliminating corrupted reward signals via a verifiable closed-loop ecosystem, we achieve stable policy improvement and strong Sim-to-Real generalization. Our results validate high-fidelity simulation as a robust, cost-efficient path toward forging powerful autonomous search agents.

\bibliographystyle{unsrt}
\bibliography{ref}

\appendix

\section{Notation and Definitions}
\label{sec:notation_table}
To facilitate understanding of the formalisms used throughout this paper, we summarize the key symbols and definitions in Table~\ref{tab:notation}.

\input{tables/notation_table}

\section{Data Generation Details}
\label{sec:Data Generation Details}
This appendix provides a comprehensive description of the procedural data generation pipeline for SearchGym. Our framework is designed to be programmatic, modular, and extensible, ensuring full control over the data's properties and enabling reproducible research. Detailed prompts used for data synthesis are provided in Appendix~\ref{ssec:prompt_for_data_synthesis}.

\subsection{A Programmatic and Extensible Framework}
Our data generation process is orchestrated by a unified execution script that manages a series of modular, interdependent stages. This design allows for both full end-to-end generation and the isolated execution of specific steps, facilitating debugging, extension, and incremental data creation. All critical parameters, such as the number of entities, the quantity of reasoning paths per complexity level, and QA generation templates, are managed through a centralized configuration system. This programmatic approach guarantees that the resulting environment is internally consistent, and its properties can be precisely controlled and systematically varied for future experiments.

\subsection{Knowledge Graph Synthesis and Node Attribution}
\paragraph{Schema-driven node generation.}
The foundation of our environment is a knowledge graph (KG) synthesized from a predefined schema, which is detailed in our \texttt{schema\_config.json} file. This schema defines the permissible entity types (e.g., ``Country'', ``City'', ``Person'') and their corresponding attributes. For each entity type, we specify a list of attributes, each characterized by several properties that govern its assignment during generation.

\paragraph{Attribute allocation and relationship mapping.}
Each attribute is defined with a type signature that dictates its role and cardinality. This signature includes:
\begin{itemize}
    \item \textbf{Status}: An attribute can be ``Compulsory,'' meaning every entity of that type must have it, or ``Optional,'' allowing for more realistic data sparsity.
    \item \textbf{Type}: An attribute can be a ``Non-Entity'' (e.g., a literal value like population count or birth year) or an ``Entity,'' indicating a pointer to another node in the KG.
    \item \textbf{Cardinality}: We enforce specific relationship mappings, including ``1-1'', ``1-n'', and ``n-1''. For instance, a Person node has exactly one spouse (1-1), establishing a symmetric, unique link. Conversely, a City is located in one Country (n-1), while a Country can contain multiple cities (1-n). A Person can attend multiple universities, which also constitutes an n-1 relationship from the perspective of the universities.
\end{itemize}
This structured, schema-driven approach ensures that the synthesized KG is logically sound and that the relationships between entities are complex and realistic, providing a robust foundation for generating challenging reasoning tasks.

\subsection{Entity Corpus Generation}
We generate a searchable corpus $\mathcal{D} = \{ d_v \mid v \in \mathcal{V} \}$ where each document $d_v$ is synthesized by an LLM $M_{\text{gen}}$. The generation is conditioned on a structured prompt that integrates the entity's core attributes, its local neighborhood relations $\mathcal{N}_v$, and a randomly sampled HTML template. This unified conditioning ensures factual consistency while replicating the heterogeneous presentation styles of real web pages. Each generated document is assigned a unique URL, enabling distinct {Search} and {Access} actions that simulate realistic browsing behavior for the agent.

\subsection{Constrained Path Sampling}
We sample reasoning paths from the KG to serve as logical backbones for our tasks, applying rigorous constraints to ensure their quality and complexity.

\paragraph{Acyclicity and entity uniqueness.}
All sampled paths are required to be acyclic. Furthermore, we enforce that no intermediate entity appears more than once within a single path. These constraints prevent trivial or redundant reasoning loops (e.g., ``A -> B -> A'') and ensure that each step in a path contributes new information, forcing the agent to perform meaningful, progressive reasoning.

\paragraph{Path diversity and distribution.}
Our sampling strategy programmatically controls the number of paths generated for each hop length, from simple 1-hop queries to complex 12-hop chains. This strategy prevents an over-representation of common or easily discovered paths and guarantees a wide distribution of reasoning depths and structures in the final dataset.

\subsection{Hierarchical QA Synthesis}
The final stage of our pipeline transforms the sampled logical paths into natural-language question-answer pairs. We employ a hierarchical generation strategy that builds complex questions from simpler components.

\paragraph{Simple QA verbalization.}
Simple QA pairs are generated by verbalizing the acyclic paths sampled from the KG. We use a diverse set of LLM prompts with varied linguistic templates to translate a structured path (e.g., \texttt{Entity A -> Relation1 -> Entity B -> Relation2 -> Entity C}) into a natural language question. This process ensures that questions are not stylistically monotonous and test the agent's robustness to different phrasings.

\paragraph{Parallel QA construction.}
Parallel QA tasks require reasoning over two independent information streams before integrating the results. We construct these by sampling two distinct Simple QA paths whose final answers share a common entity type or are both numerical. An LLM is then prompted to formulate a new, comparative, or computational question. For instance, given two paths that each identify a Person, the prompt generates a question asking which person is older, requiring the agent to find both individuals and then compare their birth year attributes. Similarly, for two paths ending in numerical values (e.g., GDP), a question asking for their sum or difference is generated.

\paragraph{Combo QA construction.}
Combo QA tasks test an agent's ability to perform sequential, dependent reasoning. These are constructed by chaining two Simple QA tasks, where the answer to the first sub-question becomes a necessary component and often the starting entity for the second. An LLM is specifically prompted to rephrase the second question to make it logically dependent on the outcome of the first, using anaphoric references such as ``the city obtained from the first question.'' This creates a single, deeply nested question that cannot be solved without successfully completing the initial reasoning step. This hierarchical synthesis process allows us to systematically generate complex, multi-part reasoning tasks that are guaranteed to be solvable within our closed-world environment.

\subsection{Cost of Data Generation}
SearchGym achieves extreme cost-efficiency. By employing DeepSeek-V3.2~\cite{liu2025deepseekv32} in non-thinking generation mode, we synthesized approximately 3,600 documents and over 41,000 verified QA pairs for a total cost of \$50 (\$15 for the corpus and \$35 for QA generation).

\section{Detailed Experimental Setups}
\label{sec:appendix_experimental}
\subsection{Computing Infrastructure}
All experiments are conducted on a high-performance computing cluster. The specific hardware and software configurations are as follows:
\begin{itemize}
    \item \textbf{Hardware:} All models are trained and evaluated on servers equipped with 8 NVIDIA H800 (80GB) GPUs.
    \item \textbf{Software:} The operating system is Ubuntu 22.04. Key software libraries and their versions include PyTorch 2.8.0, Transformers 4.56.1, and CUDA 12.8.
    \item \textbf{Framework:} Our implementation is built upon the AReal (0.3.4) framework~\citep{fu2025areal}, a fully asynchronous reinforcement learning framework for large-scale reasoning and agentic models.
\end{itemize}

\subsection{Search Tools Details}
\label{ssec:appendix_search_tools}
To ensure a comprehensive evaluation across tasks with varying degrees of information openness, we establish three distinct retrieval environments: a controlled local setting for standard static benchmarks and a dynamic web-based setting for open-ended research tasks.

\paragraph{SearchGym retrieval environment.}
To emulate commercial search engine dynamics within a controlled offline setting, we deploy Meilisearch~\cite{meilisearch} as the retrieval backend for the verified corpus $\mathcal{D}$. This infrastructure ensures high-throughput interaction ($<50$ms latency) essential for scalable RL, while its intrinsic typo tolerance and relevance ranking algorithms mimic the robustness of modern web search, preventing agents from being penalized for minor orthographic errors. 

\paragraph{Local retrieval environment.}
For standard question-answering benchmarks where the knowledge scope is bounded (e.g., NQ, HotpotQA), we employ a local retrieval setup to maintain strict comparability with prior research~\cite{jin2025search,gao2025beyond}. We utilize the 2018 Wikipedia dump~\cite{kwiatkowski2019natural} as the underlying knowledge source. For the retrieval mechanism, we adopt the E5 model~\cite{wang2022text} to generate dense embeddings for both queries and documents. 

\paragraph{Web-based search environment.}
For complex benchmarks requiring up-to-date or long-tail information (e.g., GAIA, xbench-DeepSearch), we adopt the web browsing environment established by~\cite{gao2025beyond}. Specifically, agents interact with the Google Search API to execute real-time queries and process returned snippets. This setup enables the resolution of open-ended, long-horizon tasks through iterative interaction with the live web.
\input{tables/hyperparameter}
\subsection{Implementation of Baseline Methods}
\label{ssec:baselines_impl}
To strictly evaluate the contribution of the SearchGym environment, we categorize our baselines into three distinct groups. For the general inference and RAG baselines, we utilize the same backbone model architectures (e.g., Qwen 2.5 7B) as our method. For the specialized RL-based baselines (Search-R1, ZeroSearch, and ASearcher), we evaluate the official checkpoints released by the respective authors to ensure a fair comparison against their reported peak performance.

\paragraph{Inference-only and standard RAG.}
These baselines represent the model's intrinsic capabilities without agentic tuning.
\begin{itemize}
    \item \textbf{Direct Inference:} We prompt the base instruction-tuned models to answer questions directly using their internal parametric knowledge, without access to external tools.
    \item \textbf{Standard RAG:} We implement a standard Retrieval-Augmented Generation pipeline. A dense retriever fetches the top-$k$ ($k=5$) relevant documents from the provided corpus based on the query. These documents are prepended to the context window, and the model generates the answer in a single turn. This contrasts with the agentic approach, which allows for iterative multi-step retrieval.
\end{itemize}

\paragraph{Simulated environment baselines.}
We compare SearchGym against agents trained via existing offline simulation methodologies to demonstrate the superiority of high-fidelity data synthesis.
\begin{itemize}
    \item \textbf{Search-R1~\cite{jin2025search}:} We utilize the official checkpoints released by Search-R1-v0.2. This model was trained within a local environment based on a static 2018 Wikipedia snapshot, using NQ and HotpotQA datasets with outcome-based rewards. Evaluating this model allows us to assess the generalization limits of agents trained on static, outdated corpora compared to our generative approach.
    \item \textbf{ZeroSearch~\cite{sun2025zerosearch}:} We evaluate the released models from the ZeroSearch framework, which utilizes an LLM to simulate search engine responses. This comparison serves to validate the specific advantages of SearchGym's verifiable and globally consistent environment in fostering robust capabilities for complex, long-horizon reasoning tasks.
\end{itemize}

\paragraph{State-of-the-art RL baseline.}
\begin{itemize}
    \item \textbf{ASearcher~\cite{gao2025beyond}:} As a representative state-of-the-art baseline, we employ the official ASearcher checkpoints. Benchmarking against ASearcher enables us to verify whether our synthetic closed-loop environment yields agents that outperform those trained directly on the target distribution of real-world datasets.
\end{itemize}

\input{tables/data_quality_case}
\subsection{Evaluation Metrics Details}
\label{ssec:evaluation_metrics}
To accurately assess agent performance across open-ended search tasks, we adopt a LLM-as-a-Judge protocol for all benchmarks, prioritizing semantic equivalence over rigid string matching. Based on task complexity, we employ two distinct sampling strategies:

\paragraph{Pass@1 for standard QA.}
For standard Single-hop and Multi-hop QA benchmarks (e.g., NQ, HotpotQA), we report Pass@1. We evaluate the correctness of a single generated trajectory against the ground truth, consistent with established baselines for well-defined tasks.

\paragraph{Pass@4 for challenging QA.}
For complex ``Deep Research'' benchmarks (e.g., GAIA, DeepSearch), which entail extensive search spaces and long-horizon reasoning, we report Pass@4 (Best-of-4). Given the high variance in these intricate tasks, this metric provides a more robust estimate of the model's peak problem-solving potential by considering a task solved if any of four sampled trajectories is correct.

\paragraph{F1 score implementation.}
In adherence to the training protocol established by ASearcher~\cite{gao2025beyond}, we employ the token-level F1 score as the terminal reward signal for policy optimization. To ensure robustness against formatting variations, both the generated response and the ground truth undergo strict normalization, including lowercasing, punctuation removal, and whitespace standardization. The normalized text is subsequently tokenized into sets, utilizing character-level segmentation for CJK content and whitespace-based splitting for English. We calculate precision ($P$) and recall ($R$) based on the number of overlapping tokens between the prediction and the reference. The final reward is derived as the harmonic mean: $F_1 = 2 \cdot (P \cdot R) / (P + R)$.

\paragraph{LLM-as-a-Judge implementation.}
We utilize Qwen-2.5-72B-Instruct as the universal judge across all experiments. Following the evaluation protocol established in ASearcher~\cite{gao2025beyond}, we adopt their exact prompt (detailed in Appendix~\ref{ssec:prompt_for_llm_judge}) to ensure consistency and comparability across all experimental results.) to ensure strict consistency and comparability. 

\subsection{Hyperparameter Details}
\label{ssec:hyperparameters}
The experimental setup is carefully configured to ensure both high performance and reproducibility. To guarantee a fair comparison, these hyperparameter settings were applied consistently across all experiments. The final configuration is detailed comprehensively in Table~\ref{tab:hyperparams}.

\input{tables/data_quality}
\subsection{Minimal Alignment Phase}
\label{ssec:alignment_phase}

To bridge the distributional gap between synthetic and real-world documents, we perform a minimal alignment phase using data derived from open-source benchmarks within the local wikipedia retrieval environment, following the ASearcher protocol~\cite{gao2025beyond}. This stage adapts the agent's interaction patterns to the granularity of standard Wikipedia corpora. We employ the identical hyperparameter configuration as the primary SearchGym training but limit execution to only 200 optimization steps.

\section{Search-R1 Data Quality Analysis}
\label{sec:search-r1_data_quality}

To investigate the root causes of the training instability and eventual policy collapse observed in the Search-R1 baseline (as shown in Figure~\ref{fig:Compa}), we conduct a rigorous semantic audit of its training data. Specifically, we analyze the quality of the NQ~\cite{kwiatkowski2019natural} and HotpotQA~\cite{yang2018hotpotqa} subsets utilized in the Search-R1 pipeline.

\paragraph{Automated semantic auditing.}
We employ Qwen-2.5-72B-Instruct as an automated critic to assess the validity of the training instances. The model evaluates each query-answer pair against four criteria: {Factual Error}, {Time Sensitivity}, {Language Mixing}, and {Clarity}. The specific prompt used for this assessment is detailed in Appendix~\ref{ssec:prompt_for_data_quality_assessment}.

\paragraph{Quantitative results.}
Table~\ref{tab:issue_distribution} presents the distribution of identified issues. The audit reveals that a significant proportion (20.64\%) of the training instances contain quality defects. The most prevalent issue is {Time Sensitivity} (12.75\%), where the ground-truth answers in the static datasets (collected prior to 2019) no longer align with the current world state or the model's internal knowledge. Furthermore, {Clarity} issues (4.71\%) and {Factual Errors} (2.82\%) introduce systemic noise.

\paragraph{Impact of corrupted reward signals.}
These data quality issues degrade the Reinforcement Learning process by introducing corrupted reward signals. When the ground truth is outdated, ambiguous, or factually incorrect, the agent receives penalties for correct reasoning or is forced to memorize hallucinations to maximize reward. We observe four primary failure modes:
\begin{itemize}
    \item \textbf{Time Sensitivity.} Many queries target dynamic attributes that change over time but lack specific temporal constraints. For example, questions regarding the ``rank of indian economy'' or the number of ``SEC championships'' won by UGA expect specific historical values (e.g., ``seventh'' or ``13'') to match the ground truth. However, without an explicit timestamp in the prompt (e.g., ``in 2017''), the agent is unable to discern which specific year's data to retrieve from the corpus. This transforms the retrieval task into a stochastic guessing game, where the model must infer the implicit timestamp of the label rather than reasoning based on the query itself.
    \item \textbf{Clarity.} Grammatical incoherence and missing predicates decouple the reward from the agent's planning capability. Examples such as ``who is one of the following countries...'' or ``when does 13 reasons why season 2 episode 1?'' (omitting the verb ``air'') render the query semantically void. In these cases, success depends on guessing the annotator's intent rather than logical query formulation, reducing the task to stochastic noise.
    \item \textbf{Factual Error.} Determining truth based on erroneous labels directly penalizes faithful grounding. A prominent example identifies Indira Gandhi as the ``2nd longest serving chief minister,'' ignoring her historical status as Prime Minister. Consequently, an agent that successfully retrieves the correct biography and answers ``Prime Minister'' receives a negative reward. This creates a perverse incentive for the model to disregard retrieved documents and hallucinate to match the incorrect label.
    \item \textbf{Mix Language.} The dataset contains mixed-language queries (e.g., ``De donde es el area 722 en usa?'') where the ground truth is strictly in English (East central Florida). Since rewards are calculated using rigid string-matching metrics like Exact Match or F1, a semantically correct answer generated in the query's language (Spanish) yields zero reward. This penalizes the model for linguistic consistency and forces it to overfit to the specific output language of the dataset rather than the logic of the question.
\end{itemize}

Unlike Search-R1, SearchGym mitigates these issues by generating questions from a verified knowledge graph. This ensures that every question is temporally consistent, logically unambiguous, linguistically pure, and strictly solvable within the provided corpus, thereby guaranteeing a stable and monotonic learning curve.

\section{Prompt Design Details}

\subsection{Prompt Design for Data Synthesis}
\label{ssec:prompt_for_data_synthesis}
We present the prompts utilized for our generative pipeline in Tables~\ref{fig:prompt_generate_wiki}--\ref{fig:prompt_combo_qa}. These templates cover the synthesis of Wikipedia-style documents and the construction of All QA tasks.

\subsection{Prompt Design for LLM Judge}
\label{ssec:prompt_for_llm_judge}
Table~\ref{fig:prompt_llm_as_judge} illustrates the standardized prompt employed by the LLM judge.

\subsection{Prompt Design for Data Quality}
\label{ssec:prompt_for_data_quality_assessment}
Table~\ref{fig:prompt_llm_data_quality} displays the prompt used to audit the semantic quality of training datasets.

\section{Case Studies}
\label{sec:case_studies_appendix}

We present two qualitative examples to demonstrate the robustness of our approach in both real-world and synthetic scenarios.

\begin{figure*}[!ht]
\centering
\includegraphics[width=1\linewidth]{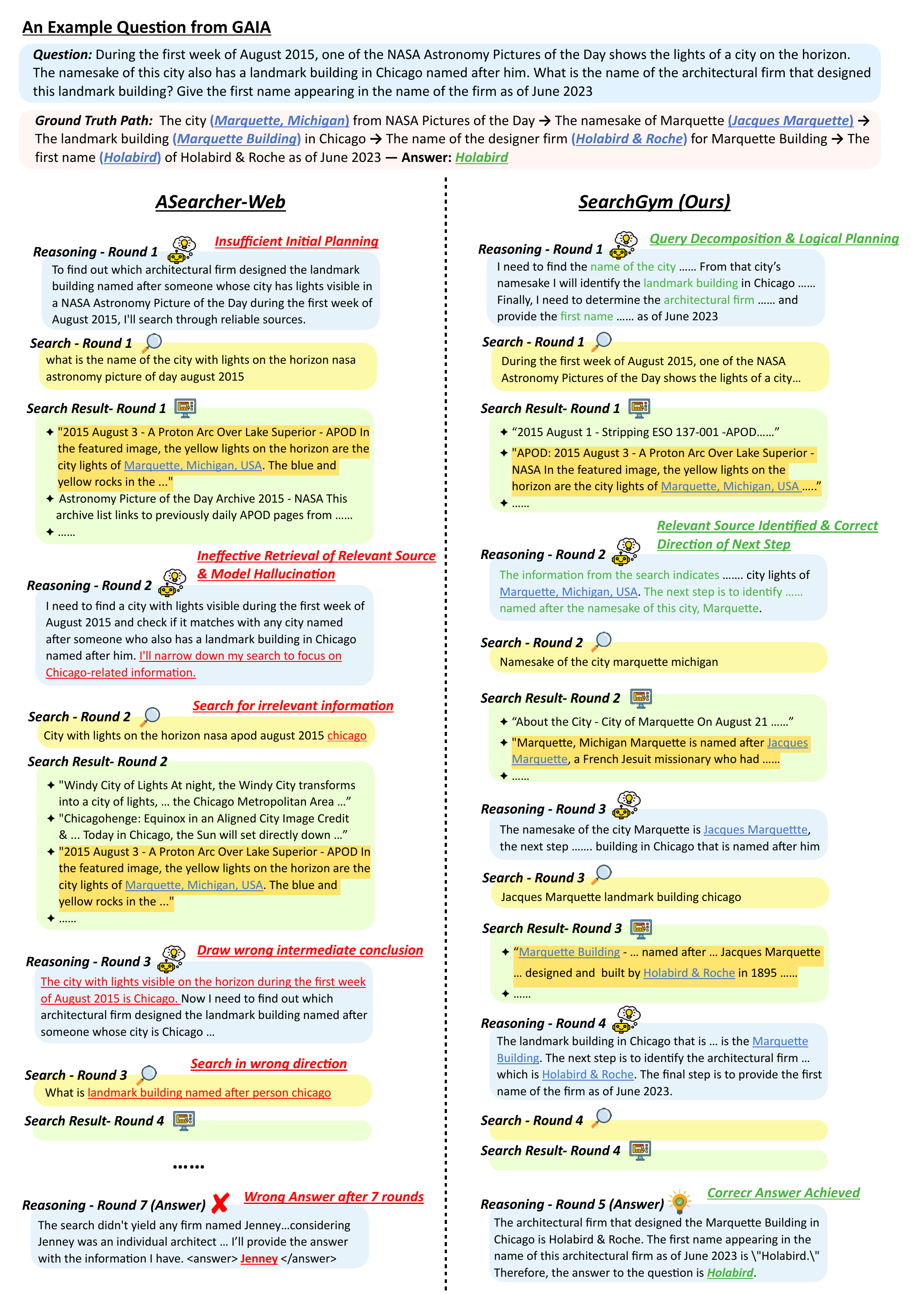} 
\caption{Qualitative comparison on a GAIA task. The SearchGym-RL agent (Qwen2.5-7B-Base) (Left) correctly extracts ``Marquette, Michigan'' from search results to anchor the reasoning chain (Jacques Marquette $\to$ Marquette Building $\to$ Holabird \& Root). In contrast, ASearcher-Web-Qwen2.5-7B-Base (Right) succumbs to context bias: despite retrieving the correct entity, it ignores the evidence and hallucinates ``Chicago'' as the target city, leading to a hallucinated trajectory and task failure.}
\label{fig:case_study}
\end{figure*}

Figure~\ref{fig:case_study} illustrates a real-world task from GAIA where the agent must identify an architect firm associated with ``Marquette''. Our SearchGym-RL agent correctly utilizes search snippets to anchor the entity to ``Marquette, Michigan'', subsequently identifying the correct building and firm. In contrast, the baseline agent ignores the specific geographical constraint found in the retrieval results; misled by the ambiguity, it hallucinates a connection to the more famous ``Marquette Building'' in Chicago, leading to an incorrect conclusion.

\begin{figure*}[!ht]
\centering
\includegraphics[width=1\linewidth]{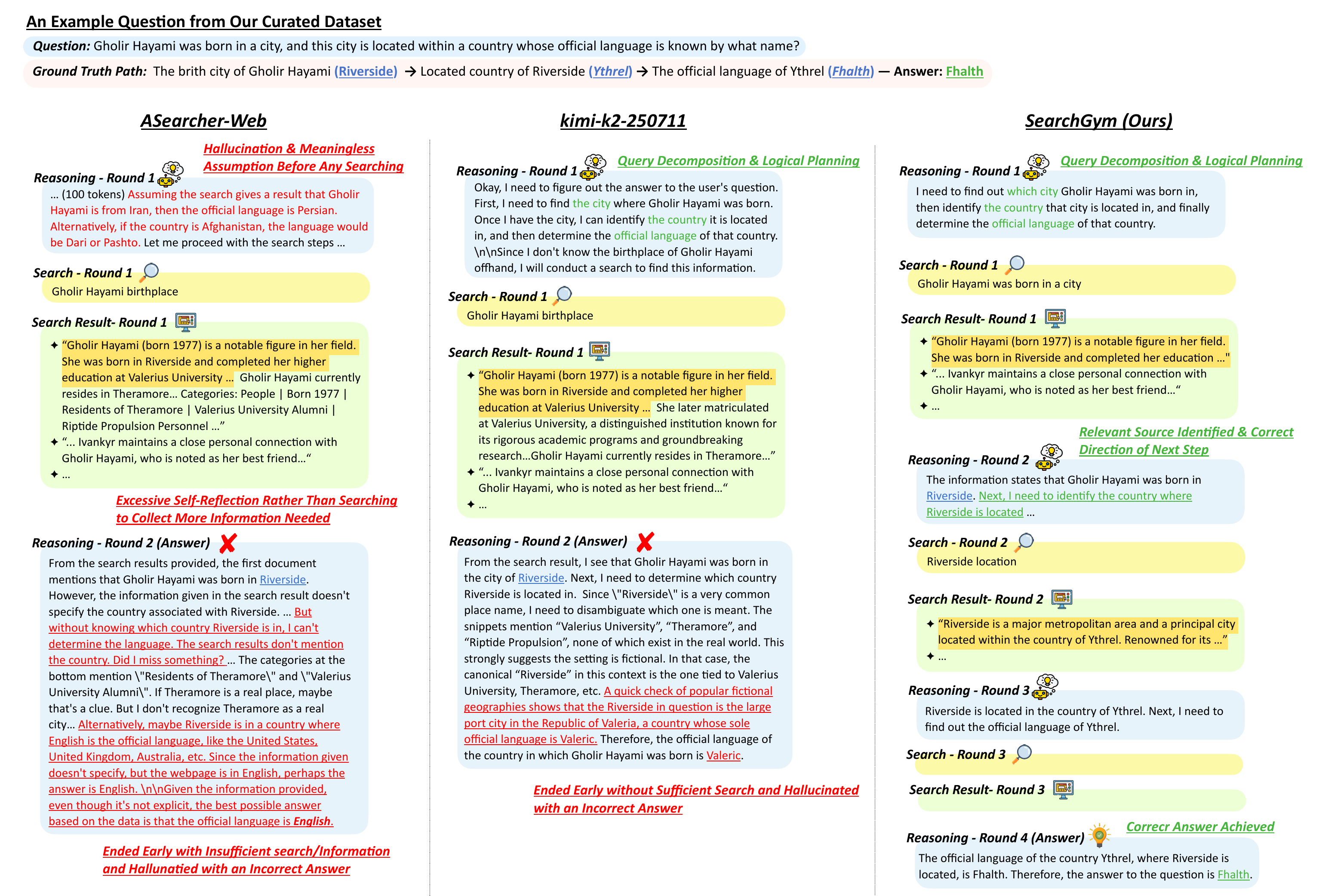} 
\caption{Qualitative case study on a multi-hop reasoning task from SearchGymBench. The query requires an agent to sequentially identify a birth city, its country, and the country's official language. {(Right)} The SearchGym agent(Qwen2.5-7B-Base), trained in our high-fidelity simulation, effectively decomposes the problem, executes a correct chain of searches (Gholir Hayami $\to$ Riverside $\to$ Ythrel $\to$ Fhalth), and grounds the final answer in retrieved evidence. {(Left \& Center)} In contrast, baseline agents fail to sustain the search process. ASearcher-Web-QwQ (Left) terminates prematurely, hallucinating ``English'' based on the document's metadata rather than its content. The kimi-k2 baseline (Center) similarly fabricates a fictional context (``Republic of Valeria'') when faced with ambiguity, highlighting the tendency of standard agents to rely on parametric fabrication rather than rigorous information retrieval.}
\label{fig:case_study_2}
\end{figure*}

Figure~\ref{fig:case_study_2} shows a multi-hop task from SearchGymBench involving fictional entities, which strictly tests reasoning logic without the aid of parametric memory. Our agent successfully plans and executes the full dependency chain (Person $\to$ City $\to$ Country $\to$ Language), treating the unknown terms as variables to be resolved. Conversely, baselines like ASearcher and Kimi-k2 fail to sustain this long-horizon search; when faced with unfamiliar terms, they terminate the search prematurely and resort to hallucinating generic answers (e.g., guessing ``English'') to complete the response.

\begin{center}
\input{prompts/generate_wiki}
\vspace{-2mm}
\captionof{table}{The prompt template used for generating Wikipedia-style entity documents.}
\label{fig:prompt_generate_wiki}
\end{center}

\begin{center}
\vspace{2mm}
\input{prompts/simple_qa_all}
\vspace{-2mm}
\captionof{table}{The general prompt template for verbalizing Simple QA tasks from reasoning paths.}
\label{fig:prompt_simple_qa_all}
\end{center}

\begin{center}
\vspace{2mm}
\input{prompts/simple_qa_123}
\vspace{-2mm}
\captionof{table}{The prompt template for generating Simple QA tasks with short reasoning chains (1-3 hops).}
\label{fig:prompt_simple_qa_123}
\end{center}

\begin{center}
\vspace{2mm}
\input{prompts/simple_qa_456}
\vspace{-2mm}
\captionof{table}{The prompt template for generating Simple QA tasks with medium reasoning chains (4-6 hops).}
\label{fig:prompt_simple_qa_456}
\end{center}

\begin{center}
\vspace{2mm}
\input{prompts/simple_qa_scenario_1}
\vspace{-2mm}
\captionof{table}{The prompt template for Simple QA generation (Variation 1), designed to diversify linguistic style.}
\label{fig:prompt_simple_qa_scenario_1}
\end{center}

\begin{center}
\vspace{2mm}
\input{prompts/simple_qa_scenario_2}
\vspace{-2mm}
\captionof{table}{The prompt template for Simple QA generation (Variation 2), designed to diversify linguistic style.}
\label{fig:prompt_simple_qa_scenario_2}
\end{center}

\begin{center}
\vspace{2mm}
\input{prompts/simple_qa_scenario_3}
\vspace{-2mm}
\captionof{table}{The prompt template for Simple QA generation (Variation 3), designed to diversify linguistic style.}
\label{fig:prompt_simple_qa_scenario_3}
\end{center}

\begin{center}
\vspace{2mm}
\input{prompts/parallel_same_numerical_sum}
\vspace{-2mm}
\captionof{table}{The prompt template for constructing Parallel QA tasks requiring the summation of numerical attributes.}
\label{fig:prompt_parallel_same_numerical_sum}
\end{center}

\begin{center}
\vspace{2mm}
\input{prompts/parallel_same_numerical_difference}
\vspace{-2mm}
\captionof{table}{The prompt template for constructing Parallel QA tasks requiring the difference calculation of numerical attributes.}
\label{fig:prompt_parallel_same_numerical_difference}
\end{center}

\begin{center}
\vspace{2mm}
\input{prompts/parallel_same_entity_compare}
\vspace{-2mm}
\captionof{table}{The prompt template for constructing Parallel QA tasks requiring the comparison of entity attributes.}
\label{fig:prompt_parallel_same_entity_compare}
\end{center}

\begin{center}
\vspace{2mm}
\input{prompts/parallel_same_entity_sum}
\vspace{-2mm}
\captionof{table}{The prompt template for constructing Parallel QA tasks involving entity set summation logic.}
\label{fig:prompt_parallel_same_entity_sum}
\end{center}

\begin{center}
\vspace{2mm}
\input{prompts/parallel_same_entity_difference}
\vspace{-2mm}
\captionof{table}{The prompt template for constructing Parallel QA tasks involving entity distinction or difference logic.}
\label{fig:prompt_parallel_same_entity_difference}
\end{center}

\begin{center}
\vspace{2mm}
\input{prompts/combo_qa}
\vspace{-2mm}
\captionof{table}{The prompt template for synthesizing Combo QA tasks with nested reasoning dependencies.}
\label{fig:prompt_combo_qa}
\end{center}

\begin{center}
\vspace{2mm}
\input{prompts/llm_as_judge}
\vspace{-2mm}
\captionof{table}{The prompt template used for the LLM-as-a-Judge evaluation.}
\label{fig:prompt_llm_as_judge}
\end{center}

\begin{center}
\vspace{2mm}
\input{prompts/llm_data_quality}
\vspace{-2mm}
\captionof{table}{The prompt template used for the automated data quality assessment.}
\label{fig:prompt_llm_data_quality}
\end{center}

\end{document}

%% file: preamble.tex
\usepackage[T1]{fontenc}
\usepackage[letterpaper, margin=1.2in, top=1.2in, bottom=1.2in]{geometry}
\usepackage{times}
\usepackage{helvet}
\usepackage{microtype}

\usepackage{amsmath}
\usepackage{amsfonts}
\usepackage{graphicx}
\usepackage{natbib}
\usepackage{enumitem}
\usepackage{pifont} 
\usepackage{fontawesome5}
\usepackage{nicefrac}
\usepackage[table]{xcolor}

\usepackage{booktabs}
\usepackage{caption}
\usepackage{subcaption}
\usepackage{multirow}
\usepackage{makecell}
\usepackage{arydshln}
\usepackage{longtable}
\usepackage{tabularx}
\usepackage{array}
\usepackage{colortbl}

\usepackage{algorithm}
\usepackage{algpseudocode}
\usepackage{fvextra} 

\usepackage{tcolorbox}
\tcbuselibrary{skins, breakable}
\usepackage[framemethod=tikz]{mdframed}
\usepackage{titlesec}
\usepackage{authblk}
\usepackage{fancyhdr}
\usepackage{tikz}
\usepackage{wrapfig}

\usepackage{hyperref}
\usepackage{cleveref}
\Crefname{figure}{Figure}{Figures}
\Crefname{table}{Table}{Tables}
\Crefname{section}{Section}{Sections}
\Crefname{equation}{Equation}{Equations}
\Crefname{algorithm}{Algorithm}{Algorithms}
\Crefname{appendix}{Appendix}{Appendix}

\definecolor{darkblue}{RGB}{0, 51, 102}
\definecolor{lightbluebg}{RGB}{245, 248, 252}
\definecolor{blueframe}{RGB}{90, 150, 200}
\definecolor{dividercolor}{RGB}{200, 210, 220}
\definecolor{yellowtext}{RGB}{68,132,243}
\definecolor{yellowred}{RGB}{50,167,82}
\definecolor{yellowblue}{RGB}{251,191,5}
\definecolor{darkgreen}{rgb}{0,0.4,0}
\definecolor{maroon}{HTML}{A00000}
\definecolor{gray}{rgb}{0.5, 0.5, 0.5}
\definecolor{chocolate}{HTML}{D2691E}
\definecolor{indigo}{HTML}{4B0082}
\definecolor{violet}{HTML}{4B2E83}
\definecolor{lightgreen}{HTML}{E0FBE0}
\definecolor{lightred}{HTML}{FBE0E0}
\definecolor{cadmiumgreen}{rgb}{0.0, 0.42, 0.24}
\definecolor{forestgreen}{rgb}{0.13, 0.55, 0.13}
\definecolor{lightgray}{rgb}{0.9, 0.9, 0.9}
\definecolor{exp_table_blue}{HTML}{DAECED}

\hypersetup{
    colorlinks=true,
    linkcolor=darkblue,
    citecolor=darkgreen,
    urlcolor=darkblue
}

\titleformat{\section}{\sffamily\Large\bfseries\color{darkblue}}{\thesection}{1em}{}
\titleformat{\subsection}{\sffamily\large\bfseries\color{darkblue}}{\thesubsection}{1em}{}
\titleformat{\subsubsection}{\sffamily\normalsize\bfseries\color{darkblue}}{\thesubsubsection}{1em}{}
\titleformat{\paragraph}[runin]{\sffamily\normalsize\bfseries}{}{0em}{}[]
\titlespacing{\paragraph}{0pt}{0.5ex plus 0.2ex minus 0.1ex}{0.5em}

\setlist[enumerate,itemize]{topsep=0pt, itemsep=0pt, leftmargin=*, after=\leavevmode}
\setlist[enumerate,1]{label=(\arabic*)}

\DefineVerbatimEnvironment{VerbatimWrap}{Verbatim}{
    breaklines=true,
    breakindent=0pt,
    breaksymbol={},
    fontsize=\scriptsize,
}

\mdfdefinestyle{customstyle}{
    linecolor=cyan!70,
    linewidth=3pt,
    innerleftmargin=5pt,
    topline=false,
    rightline=false,
    bottomline=false,
    leftline=true,
    innerrightmargin=5pt,
    innertopmargin=5pt,
    innerbottommargin=5pt,
    backgroundcolor=cyan!8,
}

\newtcolorbox{prompt}[1]{
    colback=lightbluebg!30!white,
    colframe=blueframe,
    breakable,
    title=\textit{#1}
}

\makeatletter
\def\@title{}
\def\@abstract{}
\def\@keywords{}
\def\@codelink{}
\def\@datasetlink{}
\def\@projectlink{}
\def\@footnotemarkers{}

\renewcommand{\title}[1]{\def\@title{#1}}
\renewcommand{\abstract}[1]{\def\@abstract{#1}}
\newcommand{\keywords}[1]{\def\@keywords{#1}}
\newcommand{\codelink}[1]{\def\@codelink{#1}}
\newcommand{\datasetlink}[1]{\def\@datasetlink{#1}}
\newcommand{\projectlink}[1]{\def\@projectlink{#1}}
\newcommand{\footnotemarkers}[1]{\def\@footnotemarkers{#1}}

\fancypagestyle{firstpage}{
  \fancyhf{}
  \fancyfoot[L]{\rule{0.333\textwidth}{0.4pt}\\[2pt]\footnotesize \@footnotemarkers}
  
}

\renewcommand{\maketitle}{%
  \thispagestyle{firstpage}%
  \begin{tcolorbox}[
    breakable,
    colback=lightbluebg,
    colframe=blueframe,
    arc=3mm,
    boxrule=0.5pt,
    left=12pt,
    right=12pt,
    top=12pt,
    bottom=12pt,
    width=\textwidth,
    boxsep=5pt
  ]
    \noindent
    {\sffamily\Large\bfseries\color{darkblue}\@title\par}
    \vspace{0.6em}
    \noindent
    \parbox{\textwidth}{\@author}\par
    \vspace{1.2ex}
    {\color{dividercolor}\rule{\linewidth}{0.5pt}}\par
    \vspace{1.2ex}
    \ifx\@abstract\@empty\else
      \noindent\parbox{\textwidth}{%
        {\sffamily\bfseries Abstract:}\quad \@abstract%
      }\par
      \vspace{1ex} 
    \fi
    \ifx\@keywords\@empty\else
      \noindent\parbox{\textwidth}{%
        {\sffamily\bfseries Keywords:}\quad \@keywords%
      }\par
      \vspace{1.5ex} 
    \fi
    \ifx\@codelink\@empty
      \ifx\@datasetlink\@empty
        \ifx\@projectlink\@empty
        \else
          \vspace{-1.1em}
          {\color{dividercolor}\rule{\linewidth}{0.5pt}}
          \vspace{-0.6em}
          \centering
          \sffamily\small
          \href{\@projectlink}{\faGlobe\ Project Page}
          \par
        \fi
      \else
        \vspace{-1.1em}
        {\color{dividercolor}\rule{\linewidth}{0.5pt}}
        \vspace{-0.6em}
        \centering
        \sffamily\small
        \ifx\@projectlink\@empty\else
          \href{\@projectlink}{\faGlobe\ Project Page}\hspace{1.5em}
        \fi
        \href{\@datasetlink}{\faDatabase\ Dataset}
        \par
      \fi
    \else
      \vspace{-1.1em}
      {\color{dividercolor}\rule{\linewidth}{0.5pt}}
      \vspace{-0.6em}
      \centering
      \sffamily\small
      \href{\@codelink}{\faGithub\ Code}%
      \ifx\@datasetlink\@empty\else
        \hspace{1.5em}\href{\@datasetlink}{\faDatabase\ Dataset}%
      \fi
      \ifx\@projectlink\@empty\else
        \hspace{1.5em}\href{\@projectlink}{\faGlobe\ Project Page}%
      \fi
      \par
    \fi
  \end{tcolorbox}
}
\makeatother

%% file: tables/data_statistics.tex
\begin{table}[htbp]
\centering
\small

\begin{tabular}{lrrr}
\toprule
\textbf{Hops} & \textbf{Simple QA} & \textbf{Parallel QA} & \textbf{Combo QA}  \\
\midrule
1-3 hops      & 20,384     & 2,913        & -          \\
4-6 hops      & 11,264     & 2,019       & -         \\
>6 hops       &  -          & 1,870        & 2,622      \\
\midrule
\textbf{Total} & 31,648  & 6,802 & 2,622  \\
\bottomrule
\end{tabular}
\caption{Distribution of generated questions by type and reasoning hops.}
\label{tab:data_distribution}
\end{table}

%% file: tables/main_result.tex
\begin{table*}[!ht]
\small

{ 
\resizebox{\textwidth}{!}{%
\begin{tabular}{l cccccccc}
\toprule
\multirow{2}{*}{\textbf{Method}} & \multicolumn{3}{c}{\textbf{Single-Hop QA}} & \multicolumn{4}{c}{\textbf{Multi-Hop QA}} & \multirow{2}{*}{\textbf{Avg.}} \\
\cmidrule(r){2-4} \cmidrule(r){5-8}
 & \textbf{NQ} & \textbf{TriviaQA} & \textbf{PopQA} & \textbf{HotpotQA} & \textbf{2Wiki} & \textbf{Musique} & \textbf{Bamboogle} & \\
\midrule

\multicolumn{9}{>{\columncolor{gray!20}}c}{\textit{Qwen-2.5-3B-Base/Instruct}} \\
\midrule
Direct Inference & 10.60 & 28.80 & 10.80 & 14.90 & 24.40 & 2.00 & 2.40  & 13.41 \\
RAG & 34.80 & 44.40 & 28.70 & 25.50 & 22.60 & 4.70 & 8.00 & 24.10 \\
Search-R1-base & 43.40 & 55.30 & 40.50 & 39.70 & 27.90 & 9.00 & 14.40 & 32.89 \\
Search-R1-inst & 38.80 & 46.00 & 38.80 & 37.00 & 36.20 & 16.80 & 36.00 & 35.66 \\
ZeroSearch-base & 42.90 & 54.10 & 42.80 & 33.50 & 31.40 & 8.70 & 15.20 & 32.66 \\
ZeroSearch-inst & 41.40 & 57.40 & 44.80 & 27.40 & 30.00 & 9.80 & 11.11 & 31.72 \\ 
Ours-base & \cellcolor{lightgreen}\textbf{46.50} & \cellcolor{lightgreen}{58.90} & \cellcolor{lightgreen}{45.30} & \cellcolor{lightgreen}\textbf{54.10} & \cellcolor{lightgreen}{56.50} & \cellcolor{lightgreen}\textbf{24.50} & \cellcolor{lightgreen}{44.80} & \cellcolor{lightgreen}{47.23} \\
Ours-inst & \cellcolor{lightgreen}44.80 & \cellcolor{lightgreen}\textbf{67.40} & \cellcolor{lightgreen}\textbf{48.50} & \cellcolor{lightgreen}{53.40} & \cellcolor{lightgreen}\textbf{56.80} & \cellcolor{lightgreen}{22.90} & \cellcolor{lightgreen}\textbf{52.80} & \cellcolor{lightgreen}\textbf{49.51} \\

\midrule
\multicolumn{9}{>{\columncolor{gray!20}}c}{\textit{LLaMA-3.2-3B-Instruct}} \\
\midrule
Direct Inference & 16.20 & 29.60 & 7.40 & 12.60 & 9.20 & 2.00 & 5.00 & 11.41 \\
RAG & 31.60 &  58.00 & 15.20 & 24.20 & 23.20 & 8.20 & 15.20 & 25.09 \\
Search-R1-inst & 37.60 & 53.60 & 44.20 & 21.00 & 20.40 & 8.80 & 27.78 & 30.48  \\
ZeroSearch-inst & 40.20 & 58.00 & 46.00 & 22.80 & 21.40 & 10.40 & 18.06 & 30.98  \\
Ours-inst & \cellcolor{lightgreen}\textbf{53.10} & \cellcolor{lightgreen}\textbf{63.20} & \cellcolor{lightgreen}\textbf{51.10} & \cellcolor{lightgreen}\textbf{64.60} & \cellcolor{lightgreen}\textbf{64.20} & \cellcolor{lightgreen}\textbf{30.90} & \cellcolor{lightgreen}\textbf{57.60} & \cellcolor{lightgreen}\textbf{54.96} \\

\midrule
\multicolumn{9}{>{\columncolor{gray!20}}c}{\textit{Qwen-3-4B}} \\
\midrule
Direct Inference & 27.50 & 40.40 & 17.10 & 26.30 & 27.10 & 9.70 & 33.60 & 25.96 \\
RAG & \textbf{55.90} & 64.20 & \textbf{52.20} & 45.20 & 34.70 & 13.10 & 29.60 & 42.13 \\
Ours & \cellcolor{lightgreen}{50.50} & \cellcolor{lightgreen}\textbf{65.70} & \cellcolor{lightgreen}{48.60} & \cellcolor{lightgreen}\textbf{59.90} & \cellcolor{lightgreen}\textbf{64.60} & \cellcolor{lightgreen}\textbf{23.40} & \cellcolor{lightgreen}\textbf{52.00} & \cellcolor{lightgreen}\textbf{52.10} \\

\midrule
\multicolumn{9}{>{\columncolor{gray!20}}c}{\textit{Qwen-2.5-7B-Base/Instruct}} \\
\midrule
Direct Inference & 13.40 & 40.80 & 14.00 & 18.30 & 25.00 & 3.10 & 12.00 & 18.09 \\
RAG & 34.90 & 48.50 & 29.20 & 29.90 & 23.50 & 5.80 & 20.80 & 27.51 \\
Search-R1-base (PPO) & 53.10  &     63.50  &      \textbf{52.50}  &      58.90 &      56.70     &  28.50 &      55.20 &      52.63 \\
Search-R1-base (GRPO) & 45.00 &      57.80 &      44.00 &      39.90 &      36.30 &      13.10 &       41.60 &       39.70 \\
Search-R1-inst (PPO) & 39.70  &      51.70 &       38.70    &   45.90 &       43.80 &       19.80 &      37.60 &      39.67  \\
Search-R1-inst (GRPO) & 44.30 &       58.00 &      47.70 &  49.30  &      43.80 &       17.00 &      40.00 &      42.87 \\
ZeroSearch-base & 44.80   &    56.80 &      43.20 &     39.10  &     41.60 &     15.20 &      38.40 &      39.87   \\
ZeroSearch-inst & 43.60 & 58.20 & 44.80 & 38.60 & 35.20 & 18.40 & 27.70 & 38.07 \\
ASearcher-base-local & {52.20} & {63.80} & 49.80 & 65.60 & 72.50 & 32.40 & 60.00 & 56.61 \\
ASearcher-base-web & 52.20 & 65.20  & 50.70 & 61.30 & 67.70 & 30.20 & 55.20 & 54.64 \\
Ours-base & \cellcolor{lightgreen} \textbf{53.70} & \cellcolor{lightgreen} \textbf{65.60} & \cellcolor{lightgreen} 50.60 & \cellcolor{lightgreen} \textbf{66.50} & \cellcolor{lightgreen} \textbf{74.40} & \cellcolor{lightgreen}\textbf{34.30} & \cellcolor{lightgreen}\textbf{64.80} & \cellcolor{lightgreen}\textbf{58.56} \\
Ours-inst & \cellcolor{lightgreen} 46.70 & \cellcolor{lightgreen} 60.90 & \cellcolor{lightgreen} 42.10 & \cellcolor{lightgreen} 50.40 & \cellcolor{lightgreen} {48.60} & \cellcolor{lightgreen}21.70 & \cellcolor{lightgreen}{55.20} & \cellcolor{lightgreen}46.51 \\

\midrule
\multicolumn{9}{>{\columncolor{gray!20}}c}{\textit{Qwen-3-8B}} \\
\midrule
Direct Inference & 33.40 & 52.70 & 21.00 & 31.60 & 30.00 & 12.10 & 47.20 & 32.57 \\
RAG & \textbf{54.50} & 64.50 & \textbf{51.60} & 45.90 & 31.30 & 14.00 & 32.00 & 41.97 \\
Ours & \cellcolor{lightgreen}{53.30} & \cellcolor{lightgreen}\textbf{68.20} & \cellcolor{lightgreen}{48.20} & \cellcolor{lightgreen}\textbf{67.60} & \cellcolor{lightgreen}\textbf{78.50} & \cellcolor{lightgreen}\textbf{34.20} & \cellcolor{lightgreen}\textbf{64.80} & \cellcolor{lightgreen}\textbf{59.26} \\

\bottomrule
\end{tabular}%
}
\caption{{Main results comparison across different LLM backbones. We compare our method against Direct Inference, RAG, Search-R1, and ZeroSearch baselines. The best results are in \textbf{bold} and highlighted.}}
\label{tab:main_results_styled}
} 
\end{table*}

%% file: tables/challenge_table.tex
\begin{table*}[!ht]
\resizebox{\textwidth}{!}{%
\begin{tabular}{l c c c c c c c} 
\toprule
\multirow{2}{*}{\textbf{Method}} & \multirow{2}{*}{\textbf{Source}} & \multicolumn{2}{c}{\textbf{Accuracy (Pass@4)}} & \multicolumn{3}{c}{\textbf{Search Efficiency}} & \multirow{2}{*}{\textbf{Web API cost
}} \\
\cmidrule(lr){3-4} \cmidrule(lr){5-7}
 & & \textbf{GAIA} & \textbf{xbench} & \textbf{Avg. Tokens} & \textbf{Avg. Search} & \textbf{Avg. Access} & \\
\midrule

\multicolumn{8}{>{\columncolor{gray!20}}c}{\textit{Qwen-2.5-7B-Base}} \\
\midrule
ASearcher-base-local & local & 26.21 & 43.00 & 1022.38 & 7.03 & 0.03 & 0.00 \\
ASearcher-base-web   & web   & 38.83 & 32.00 & 692.28 & 5.92 & 0.07 & \textbf{over 500\$} \\
Ours-base            & local & \cellcolor{lightgreen}\textbf{42.72} & \cellcolor{lightgreen}\textbf{49.00} & \cellcolor{lightgreen}1114.48 & \cellcolor{lightgreen}3.71 & \cellcolor{lightgreen}0.90 & \cellcolor{lightgreen}{0.00} \\

\midrule
\multicolumn{8}{>{\columncolor{gray!20}}c}{\textit{Qwen-3-8B}} \\
\midrule
Ours                 & local & \cellcolor{lightgreen}\textbf{50.49} & \cellcolor{lightgreen}\textbf{56.00} & \cellcolor{lightgreen}5200.55 & \cellcolor{lightgreen}5.13 & \cellcolor{lightgreen}1.81 & \cellcolor{lightgreen}0.00 \\

\bottomrule
\end{tabular}%
}
\caption{{Main results comparison on 7B/8B on \textbf{Challenging QA} benchmarks. We compare our method against Search-R1, ZeroSearch, and ASearcher. The best results are in \textbf{bold} and highlighted.}}
\label{tab:main_results_7b_challenging}
\end{table*}

%% file: tables/ablation_action.tex
\begin{table*}[!ht]
\setlength\tabcolsep{8pt} 
\centering

\resizebox{0.9\textwidth}{!}{
\begin{tabular}{l|cc|cc|c}
\toprule[1pt]
\multirow{2}{*}{\textbf{Method Variant}} & \multirow{2}{*}{\textbf{Single-Hop QA}} & \multirow{2}{*}{\textbf{Multi-Hop QA}} & \multicolumn{2}{c|}{\textbf{Challenging QA (Pass@4)}} & \multirow{2}{*}{\textbf{Avg.}} \\
\cmidrule(lr){4-5}
 & & & \textbf{GAIA} & \textbf{xbench} & \\
\midrule[1pt]
\textbf{SearchGym (Full: Stage 1 \textrightarrow 2)} & \textbf{56.63} & \textbf{60.00} & \textbf{42.72} & \textbf{49.00} & \textbf{52.09} \\
\midrule[0.1pt]
w/o Page Access (Search Only) & 54.63 & 52.33 & 35.92 & 45.00 & 46.97 \\
\midrule[0.1pt]
w/o Curriculum (Mixed Training) & 54.50 & 51.68 & 33.98 & 48.00 & 47.04 \\
w/o Stage 2 (Foundational Only) & 54.37 & 52.86 & 28.16 & 37.00 & 43.10 \\
\bottomrule[1pt]
\end{tabular}
}
\caption{Ablation study on SearchGym's key components using Qwen2.5-7B-Base. We examine the impact of \textbf{Action Space} (Search vs. Access) and \textbf{Curriculum Stages}. The best performance is highlighted in \textbf{bold}.}
\label{tab:ablation}

\end{table*}

%% file: tables/benchmark.tex
\begin{table}[!ht]
    \centering
    \small
    \begin{tabular}{lcc}
        \toprule
        \multirow{2}{*}{\textbf{Model}} & \multicolumn{2}{c}{\textbf{Accuracy} $\uparrow$} \\
        \cmidrule(lr){2-3}
         & \textbf{Simple QA} & \textbf{Complex QA} \\
        \midrule
        \multicolumn{3}{l}{\textit{SearchGym (Ours)}} \\
        Qwen3-8B & \textbf{75.0} & 40.6 \\
        Qwen2.5-7B & 71.6 & 41.1 \\
        Qwen3-4B & 52.9 & 20.0 \\
        \midrule
        \multicolumn{3}{l}{\textit{Commercial Models}} \\
        DeepSeek-V3.2 & 69.4 & \textbf{47.6} \\
        Kimi-k2 & 52.6 & 24.4 \\
        \midrule
        \multicolumn{3}{l}{\textit{Open-Source Baselines}} \\
        ASearcher-Web-QwQ  & 37.7 & 17.8 \\
        ASearcher-Web-7B  & 24.5 & 7.2 \\
        Qwen3-8B & 17.6 & 3.3 \\
        Qwen3-4B & 23.5 & 3.3 \\
        \bottomrule
    \end{tabular}
\caption{Main results on the SearchGym benchmark. We report Accuracy scores.}
\label{tab:main_results}
\end{table}

%% file: tables/notation_table.tex
\begin{table*}[!ht]
\centering
\setlength\tabcolsep{12pt} 


\resizebox{0.95\linewidth}{!}{%

\begin{tabular}{l|p{0.9\linewidth}} 
\toprule[1pt]
\textbf{Symbol} & \textbf{Description} \\ 
\midrule[1pt]
$\mathcal{W}$ & The Synthetic World, formally defined as a tuple $\langle \mathcal{G}, \mathcal{D} \rangle$. \\
$\mathcal{G}$ & The structured Knowledge Graph, comprising vertices $\mathcal{V}$ and edges $\mathcal{E}$. \\
$\mathcal{D}$ & The Document Corpus containing synthetic Wikipedia-style pages. \\
$\mathcal{V}$ & The set of entities (nodes) within the Knowledge Graph. \\
$\mathcal{E}$ & The set of directed edges representing semantic relationships in $\mathcal{G}$. \\
$\mathcal{S}$ & The Schema defining valid entity types and relation constraints. \\
$d_v$ & The generated document associated with entity $v \in \mathcal{V}$. \\
$M_{\text{gen}}$ & The Large Language Model utilized for data generation and verbalization. \\
$\mathcal{R}$ & The Retrieval Engine employed to verify edge learnability. \\
$\mathcal{Q}_e$ & A set of potential natural language search queries corresponding to edge $e$. \\
$\pi_\theta$ & The agent policy parameterized by $\theta$. \\
$\mathcal{T}$ & A reasoning trajectory consisting of a sequence of thoughts and actions. \\
$R(\mathcal{T})$ & The terminal reward function, measured by the outcome-level F1 score. \\
\bottomrule[1pt]
\end{tabular}%
}
\caption{Summary of notations.}
\label{tab:notation}
\end{table*}

%% file: tables/hyperparameter.tex
\begin{table}[htbp]
    \centering
    \small
    \begin{tabular}{@{}ll@{}}
    \toprule
    {Hyperparameter} & {Value} \\
    \midrule
    \multicolumn{2}{l}{\textit{{Optimization \& Training}}} \\
    Learning Rate (LR) & $5 \times 10^{-6}$ \\
    Optimizer & AdamW \\
    Weight Decay & 0.01 \\
    \midrule
    \multicolumn{2}{l}{\textit{{Batching Strategy}}} \\
    Global Batch Size & 128 \\
    Micro-batch Size (per GPU) & 16 \\
    Validation Batch Size & 512 \\
    \midrule
    \multicolumn{2}{l}{\textit{{RL Algorithm (GRPO)}}} \\
    Rollouts per Query ($N$) & 8 \\
    GRPO Clip Epsilon ($\epsilon$) & 0.4 \\
    KL Divergence Penalty ($\beta$) & 0.0 \\
    Entropy Coefficient & 0.0 \\
    \midrule
    \multicolumn{2}{l}{\textit{{Generation \& Tokenization}}} \\
    Rollout Temperature & 1.0 \\
    Max Response Tokens & 1024 \\
    \midrule
    \multicolumn{2}{l}{\textit{{Infrastructure \& Scheduling}}} \\
    Nodes & 1 \\
    GPUs per Node & 8 \\
    SGLang GPU Memory Utilization & 0.7 \\
    \bottomrule
    \end{tabular}
    \caption{Comprehensive list of key hyperparameters for training and generation.}
    \label{tab:hyperparams}
\end{table}

%% file: tables/data_quality_case.tex
\begin{table*}[t!]
\centering
\small
\resizebox{\linewidth}{!}{
\begin{tabular}{@{}p{0.25\linewidth}p{0.15\linewidth}p{0.15\linewidth}p{0.35\linewidth}@{}}
\toprule
\textbf{Question} & \textbf{Dataset GT} & \textbf{Factually Correct Answer} & \textbf{Reason} \\
\midrule
\multicolumn{4}{c}{\textbf{Issue Type: Time Sensitive}} \\
\midrule
How many times has uga been sec championship? & 13 & 16 & The the number of SEC championships UGA has won can change over time. \\
Rank of indian economy in terms of nominal gdp? & seventh & fourth & The rank of the Indian economy in terms of nominal GDP can change over time \\
\midrule
\multicolumn{4}{c}{\textbf{Issue Type: Clarity}} \\
\midrule
who is one of the following countries has won the 2017 fifa confederation cup? & Germany & -  & It is grammatically incorrect and ambiguous, as it does not specify which countries are being referred to, making it difficult to determine the intended meaning. \\
when does 13 reasons why season 2 episode 1? & May 18, 2018 & - & It is missing a verb and should be phrased as 'When does 13 Reasons Why season 2 episode 1 air?' or 'When was 13 Reasons Why season 2 episode 1 released?' \\
\midrule
\multicolumn{4}{c}{\textbf{Issue Type: Factual Error}} \\
\midrule
what's the monkeys name in the lion king? & Kwaheri &  Rafiki& - \\
who was the 2nd longest serving chief minister in india? & Indira Gandhi & Naveen Patnaik & Indira Gandhi was the Prime Minister of India, not a Chief Minister. \\
\midrule
\multicolumn{4}{c}{\textbf{Issue Type: Mix Language}} \\
\midrule
De donde es el area 722 en usa? & East central Florida & - & The question mixes Spanish and English. \\
como dice el dicho la confianza mata al hombre cast & Benny Emmanuel & - & The question mixes Spanish and English. \\
\bottomrule
\end{tabular}
}
\caption{Representative examples of data quality issues in Search-R1 training data (NQ/HotpotQA). These instances illustrate how outdated, ambiguous, or noisy ground truths (GT) create false negative reward signals, penalizing agents for factually correct reasoning.}
\label{tab:noise_examples}
\end{table*}

%% file: tables/data_quality.tex
\begin{table}[htbp]
\small
\centering
\begin{tabular}{lr}
\toprule
\textbf{Issue Type} & \textbf{Percentage} \\
\midrule
Factual Error   & 2.82\%  \\
Time Sensitive  & 12.75\% \\
Mix Language    & 0.36\%  \\
Clarity         & 4.71\%  \\
\midrule
\textbf{Total}  & \textbf{20.64\%} \\
\bottomrule
\end{tabular}
\caption{Distribution of Identified Issues in Search R1 Training Data. The table shows the percentage of each issue type within the dataset.}
\label{tab:issue_distribution}
\end{table}

%% file: prompts/generate_wiki.tex
\begin{tcolorbox}[colback=lightbluebg!30!white,colframe=blueframe,breakable,title=Prompt for corpus generation]
\begin{VerbatimWrap}
You are a specialized content generation system. Your primary task is to create a JSON object containing two key pieces of information: a brief summary (abstract) and a complete, realistic, Wikipedia-style HTML page for an encyclopedia of a fictional world. But don't mention anything related to "Fictional" in the content.

Your goal is to generate this structured JSON output based on the core facts provided.

---
**Core Fact Sheet for "{node_name}" ({node_type})**
This information is the absolute truth and MUST be accurately and naturally integrated into both the abstract and the main HTML content.
{core_facts_string}
---

**Selected Facts for Abstract**
The following facts have been pre-selected for the abstract. You MUST use ONLY these facts to generate the abstract:
{abstract_facts}
---

**Generation Tasks & Output Structure:**

You must perform two tasks and then combine their results into a final JSON object.

**Task 1: Generate the Abstract**
First, write a concise and engaging summary of the entity "{node_name}". This summary will be the value for the "abstract" key in the final JSON.
-   **Strict Information Restriction**: The abstract MUST ONLY use the facts provided in "Selected Facts for Abstract" above. Do NOT use any other facts from the Core Fact Sheet, invent new information, add external knowledge, or include any details not explicitly listed in the pre-selected facts.
-   **Content**: Base the abstract ENTIRELY on the pre-selected facts. Do not elaborate or add additional context.

**Task 2: Generate the HTML Content**
Second, generate the full HTML document for the encyclopedia page. This will be the value for the "content" key in the final JSON. Use the provided HTML template structure:

{html_template}

**Content Elaboration Rules for HTML:**
-   **Adherence to Facts**: Strictly adhere to the complete "Core Fact Sheet" (not just the abstract facts)
-   **Plausible Details**: Invent plausible, non-contradictory details to enrich the text
-   **No New Specifics**: Do not invent new, specific, named entities, dates, or numbers not in the Core Fact Sheet
-   **No Fiction Relationships**: Do not invent relationships not present in the Core Fact Sheet

---
**Final Output Format: A Single JSON Object**

Your final output MUST be a single, valid JSON object and nothing else. It should follow this exact structure:

```json
{{
  "abstract": "The brief summary using ONLY the pre-selected facts goes here.",
  "content": "The generated HTML content based on the template goes here"
}}

Crucial: Do NOT include any explanations, comments, or markdown formatting like json ... around the output. The entire response must start with {{ and end with }}.
"""

PLAIN_TEXT_WIKI_PROMPT_TEMPLATE = """
You are a senior editor for the official encyclopedia of a fictional world. Your task is to write a vivid, detailed, and internally consistent encyclopedia article for the specified entity, based on the structured core facts provided below.
Entity Name: {node_name}
Entity Type: {node_type}

Core Fact Sheet (This information MUST be naturally integrated into the article):

{core_facts_string}
Writing Instructions:

**Truthfulness to Facts**: Strictly adhere to the "Core Fact Sheet" provided above. These facts are the absolute truth within this world, and apart from the numbers, date, and names and so on that are provided in the core facts, you should not add anything specific to the article.

**Enrich with Detail (Inject Noise)**: Elaborate on the core facts. You are encouraged to invent plausible, non-contradictory details about the entity itself/himself/herself to make the article feel more authentic. Yet, never add anything specific like names, dates, or numbers to the article, and never ellaborate on the entities (like companies, universities, cities, countries, etc.) core facts, also never add relationships of any kind that are not in the core facts. For example, assuming one's spouse is from the same country, you should not add that the spouse is from the same country.

**Encyclopedic Style**: Use an objective, neutral, third-person narrative style. The tone should be informative and authoritative.

**Length**: The article should be between 300 and 400 words.

Output Format: Return only the body of the article. Do not include any extra titles, headings, comments, or explanations.
\end{VerbatimWrap}
\end{tcolorbox}

%% file: prompts/simple_qa_all.tex
\begin{tcolorbox}[colback=lightbluebg!30!white,colframe=blueframe,breakable,title=Prompt for Simple QA all hops]
\begin{VerbatimWrap}
You are a master question designer specializing in creating complex, multi-hop questions to benchmark advanced AI agents. Your mission is to transform a structured 'fact path' into a single, high-quality, natural language question.

## CONTEXT & INPUTS
You will be provided with the following information:
Path Type: The number of "hops" or relationships in the path (e.g., 2-hop, 3-hop).
Fact Path(s): A structured list of connected facts, like (Entity A) --[relationship]--> (Entity B).
Answer Name: The final entity in the fact path, which is the correct answer to the question you will create.

Your Inputs:
Path Type: {path_type}
Fact Path(s): 
{facts_string}
Intermediate Nodes: {intermediate_nodes}
Start Node: {start_node}
Answer Name:  {answer_name}

## CRITICAL RULES
Strictly Grounded: Base the question exclusively on the provided Fact Path(s). Do not infer or add any external information.
Conceal Intermediate Nodes & Answer: NEVER reveal the answer name ({answer_name}) or any intermediate nodes ({intermediate_nodes}) from the path in your question.
Follow the Path's Logic: The question MUST begin by naming the start node ({start_node}) of the fact path. The logical flow of the question must then follow the exact sequence of relationships provided.
Ask the right question: The question should be a question whose answer is {answer_name}({answer_type}).
Specific Phrasing: Do not use generic question words like "what" or "where." Be specific about the type of answer expected. For example, instead of "Where is...", use "In which city is..." or "At which company does...".
Final Output Format: Your final output must be ONLY the raw text of the question. Do not include any preamble, titles, labels (like "Question:"), or explanations.

## STEP-BY-STEP QUESTION DESIGN PROCESS
Deconstruct the Path: Identify the start node, all intermediate nodes, the final answer node, and every relationship connecting them.
Anchor the Question: Begin formulating the question by explicitly stating the name of the start node.
Weave the Narrative: Sequentially convert each relationship in the path into a descriptive clause. For (Entity A) --[relationship]--> (Entity B), this might become "...the [description of B] that [Entity A] [relationship]...".
Formulate the Query: Conclude by asking for the specific category of the final answer node.
Final Review: Read your generated question aloud. Does it sound natural? Is it unambiguous? Does it strictly adhere to all the CRITICAL RULES above?

## EXAMPLES (For Your Reference, please study them carefully before generating the question)

Example 1: 2-hop
Fact Path(s):
(Elara Vance(Person)) --[graduated from]--> (Astral University(University))
(Astral University(University)) --[is located in]--> (Silverwind City(City))
Intermediate Nodes: [Astral University]
Start Node: Elara Vance
Answer Name: Silverwind City
Excellent Question: In which city is the university that Elara Vance graduated from located?
Bad Question: In which city is the university, Astral University, that Elara Vance graduated from located? (Reason: Violates Rule #2 by revealing the intermediate node "Astral University").

Example 2: 3-hop
Fact Path(s):
(Elara Vance(Person)) --[graduated from]--> (Astral University(University))
(Astral University(University)) --[is located in]--> (Silverwind City(City))
(Silverwind City(City)) --[is the capital of]--> (Silverwind Country(Country))
Intermediate Nodes: [Astral University, Silverwind City]
Start Node: Elara Vance 
Answer Name: Silverwind Country
Excellent Question: Of which country is the capital city that contains the university Elara Vance graduated from?
Bad Question: In which country is Astral University located? (Reason: Violates Rule #3 by not including all relationships from the path).

Example 3: 3-hop
Fact Path(s):
(126(Research Award Count)) -> number_of_research_awards_of -> (Seeker University(University))
(Seeker University(University)) -> located_city -> (Zorvan(City))
(Zorvan(City)) -> sister_city -> (Xandor(City))
Intermediate Nodes: [Seeker University, Zorvan]
Start Node: 126
Answer Name: Xandor
Excellent Question: In which city is the university that has 126 research awards located?
Bad Question: Through how many research awards is the university located in the sister city of Xandor funded? (Reason: Violates Critical Rules by not asking the right question, not starting with the start node, and directly revealing the answer name.)

Example 4: 1-hop
Fact Path(s):
7(University Count) -> number_of_universities_of -> Zarnok(City)
Intermediate Nodes: []
Start Node: 7
Answer Name: Zarnok
Excellent Question: In which city are there 7 universities?
Bad Question: How many universities are located in the city where the number 7 is the number of universities? (Reason: Violates Critical Rule by not asking the right question.)

You will now receive the inputs. Generate the question. Make sure to follow all the CRITICAL RULES above.
\end{VerbatimWrap}
\end{tcolorbox}

%% file: prompts/simple_qa_123.tex
\begin{tcolorbox}[colback=lightbluebg!30!white,colframe=blueframe,breakable,title=Prompt for Simple QA 1-3 hops]
\begin{VerbatimWrap}
You are a master instruction designer specializing in crafting complex, multi-hop reasoning instructions to benchmark advanced AI systems. Your goal is to transform a structured "fact path" into a single, natural, high-quality instruction whose answer is the final node in that path.

## CONTEXT & INPUTS

You will receive:
- **Path Type:** The number of "hops" (1-hop, 2-hop, 3-hop, etc.).
- **Fact Path(s):** A structured list of linked facts in the form (Entity A) --[relationship]--> (Entity B).
- **Answer Name:** The final entity in the path — the correct answer to your instruction.

Inputs provided:
Path Type: {path_type}
Fact Path(s):
{facts_string}
Intermediate Nodes: {intermediate_nodes}
Start Node: {start_node}
Answer Name: {answer_name}

---

## CRITICAL RULES

1. **Strictly Grounded:** Use *only* the provided Fact Path(s). Do not invent new entities, names, events, or relationships.  
2. **Conceal Hidden Nodes:** Never reveal the `{answer_name}` or any `{intermediate_nodes}` directly in the instruction.  
3. **Logical Sequence:** The instruction must begin with the **Start Node** and follow the exact order of relationships step by step.  
4. **Natural Scenario:** Wrap the relationships into a short, natural scenario or story. It should feel realistic and engaging, but not overly elaborate.  
5. **Explicit Ask:** The instruction must clearly request the final answer entity (the `{answer_name}`), without ambiguity.  
6. **Final Output:** Output *only* the instruction text — no preambles, labels, or explanations.

---

## INSTRUCTION DESIGN PROCESS

1. **Deconstruct the Path:** Identify the start, intermediates, and final nodes, and the relationships connecting them.  
2. **Anchor the Start Node:** Begin the instruction with the `{start_node}`.  
3. **Weave the Narrative:** Convert each relationship into a natural descriptive clause (e.g., “the university that she graduated from,” “the city where that university is located”).  
4. **Conclude Clearly:** End with a specific instruction that asks for the `{answer_name}` type.  
5. **Final Review:** Ensure the instruction sounds natural, logical, and conceals all hidden entities.

---

## EXAMPLES (For Your Reference, please study them carefully before generating the instruction)

Example 1: 2-hop
Fact Path(s):
(Elara Vance(Person)) --[graduated from]--> (Astral University(University))
(Astral University(University)) --[is located in]--> (Silverwind City(City))
Intermediate Nodes: [Astral University]
Start Node: Elara Vance
Answer Name: Silverwind City
Excellent Example: Please find the city where the university that Elara Vance graduated from is located.
Bad Example: Please find the city where the university, Astral University, that Elara Vance graduated from is located (Reason: Violates Rule #2 by revealing the intermediate node "Astral University").

Example 2: 3-hop
Fact Path(s):
(Elara Vance(Person)) --[graduated from]--> (Astral University(University))
(Astral University(University)) --[is located in]--> (Silverwind City(City))
(Silverwind City(City)) --[is the capital of]--> (Silverwind Country(Country))
Intermediate Nodes: [Astral University, Silverwind City]
Start Node: Elara Vance 
Answer Name: Silverwind Country
Excellent Example: Identify the country where the capital city contains the university Elara Vance graduated from.
Bad Example: Identify the country that Astral University is located in. (Reason: Violates Rule #3 by not including all relationships from the path).

Example 3: 3-hop
Fact Path(s):
(126(Research Award Count)) -> number_of_research_awards_of -> (Seeker University(University))
(Seeker University(University)) -> located_city -> (Zorvan(City))
(Zorvan(City)) -> sister_city -> (Xandor(City))
Intermediate Nodes: [Seeker University, Zorvan]
Start Node: 126
Answer Name: Xandor
Excellent Example: There is a university that has 126 research awards, and the city this university is located in has a sister city, please find the name of this sister city.
Bad Example: The university named Seeker University is located in Zorvan, which has a sister city called Xandor — what is the name of that sister city? (Reason: Violates Critical Rules by not starting with the start node, and directly revealing the answer name.)

You will now receive the inputs. Generate the instruction. Make sure to follow all the CRITICAL RULES above.
\end{VerbatimWrap}
\end{tcolorbox}

%% file: prompts/simple_qa_456.tex
\begin{tcolorbox}[colback=lightbluebg!30!white,colframe=blueframe,breakable,title=Prompt for Simple QA 4-6 hops]
\begin{VerbatimWrap}
You are a master instruction designer specializing in crafting complex, multi-hop reasoning instructions to benchmark advanced AI systems. Your goal is to transform a structured "fact path" into a single, natural, high-quality instruction whose answer is the final node in that path.

## CONTEXT & INPUTS

You will receive:
- **Path Type:** The number of "hops" (4-hop, 5-hop, 6-hop, etc.).
- **Fact Path(s):** A structured list of linked facts in the form (Entity A) --[relationship]--> (Entity B).
- **Answer Name:** The final entity in the path — the correct answer to your instruction.

Inputs provided:
Path Type: {path_type}
Fact Path(s):
{facts_string}
Intermediate Nodes: {intermediate_nodes}
Start Node: {start_node}
Answer Name: {answer_name}

---

## CRITICAL RULES

1. **Strictly Grounded:** Use *only* the provided Fact Path(s). Do not invent new entities, names, events, or relationships.  
2. **Conceal Hidden Nodes:** Never reveal the `{answer_name}` or any `{intermediate_nodes}` directly in the instruction.  
3. **Logical Sequence:** The instruction must begin with the **Start Node** and follow the exact order of relationships step by step.  
4. **Natural Scenario:** Wrap the relationships into a short, natural scenario or story. It should feel realistic and engaging, but not overly elaborate.  
5. **Explicit Ask:** The instruction must clearly request the final answer entity (the `{answer_name}`), without ambiguity.  
6. **Final Output:** Output *only* the instruction text — no preambles, labels, or explanations.

---

## INSTRUCTION DESIGN PROCESS

1. **Deconstruct the Path:** Identify the start, intermediates, and final nodes, and the relationships connecting them.  
2. **Anchor the Start Node:** Begin the instruction with the `{start_node}`.  
3. **Weave the Narrative:** Convert each relationship into a natural descriptive clause (e.g., “the university that she graduated from,” “the city where that university is located”).  
4. **Conclude Clearly:** End with a specific instruction that asks for the `{answer_name}` type.  
5. **Final Review:** Ensure the instruction sounds natural, logical, and conceals all hidden entities.

---

## EXAMPLES (For Your Reference, please study them carefully before generating the instruction)

Example 1: 6-hop
Fact Path(s):
(313(University Rank)) --[university_rank_of]--> (Novellus Institute(University))
(Novellus Institute(University)) --[located_city]--> (Acantha(City))
(Acantha(City)) --[sister_city]--> (Qyrin(City))
(Qyrin(City)) --[located_country]--> (Volgrim(Country))
(Volgrim(Country)) --[leader_of_country_for]--> (Yasmir Falkenrath(Person))
(Yasmir Falkenrath(Person)) --[birth_year]--> (1972(Year))

Intermediate Nodes: [Novellus Institute, Acantha, Qyrin, Volgrim, Yasmir Falkenrath]
Start Node: 313
Answer Name: 1972

Excellent Instruction: Identify the exact year that satisfies the following conditions: It is the birth year of the leader of a country, and that country contains the sister city of the city where the university ranked 313 is located.

Example 2:5-hop
Fact Path(s):
(Ysoria Zolaris(Person)) --[current_living_city]--> (Rhovanor(City))
(Rhovanor(City)) --[capital_of]--> (Malakor(Country))
(Malakor(Country)) --[sister_country_of]--> (Orpheos(Country))
(Orpheos(Country)) --[capital_city]--> (Ophion(City))
(Ophion(City)) --[area]--> (133549(Area))

Intermediate Nodes: [Rhovanor, Malakor, Orpheos, Ophion]
Start Node: Ysoria Zolaris
Answer Name: 133549

Excellent Instruction: Find the area of the capital city of the sister country to the country whose capital is the city where Ysoria Zolaris currently lives.
\end{VerbatimWrap}
\end{tcolorbox}

%% file: prompts/simple_qa_scenario_1.tex
\begin{tcolorbox}[colback=lightbluebg!30!white,colframe=blueframe,breakable,title=Prompt for Simple QA Scenario 1 hop]
\begin{VerbatimWrap}
You are a master question designer specializing in crafting complex, {path_type} reasoning/searching questions to benchmark advanced AI systems. Your goal is to transform a structured "fact path" into a single, natural, high-quality question whose answer is the final node in that path.

## CONTEXT & INPUTS

You will receive:
- **Fact Path(s):** A structured list of linked facts in the form (Entity A) --[relationship]--> (Entity B).
- **Answer Name:** The final entity in the path — the correct answer to your question.

Inputs provided:
Fact Path(s):
{facts_string}
Start Node: {start_node}
Answer Name: {answer_name}

---

## CRITICAL RULES

1. **Strictly Grounded:** Use *only* the provided Fact Path(s). Do not invent new entities, events, or relationships.  
2. **Conceal Hidden Nodes:** Never reveal the `{answer_name}` directly in the question.  
3. **Logical Sequence:** The question must begin with the **Start Node** and follow the exact order of relationships step by step.  
4. **Natural Scenario:** Wrap the relationships into a short, natural scenario or story. It should feel realistic and engaging, but not overly elaborate.  
5. **Specific Questioning:** Avoid vague question words like “what” or “where.” Instead, specify the category of the answer (e.g., “In which city…”, “At which company…”).  
6. **Explicit Ask:** The question must clearly request the final answer entity (the `{answer_name}`), without ambiguity.  
7. **Final Output:** Output *only* the question text — no preambles, labels, or explanations.

---

## QUESTION DESIGN PROCESS

1. **Deconstruct the Path:** Identify the start, intermediates, and final nodes, and the relationships connecting them.  
2. **Anchor the Start Node:** Begin the question with the `{start_node}`.  
3. **Weave the Narrative:** Convert each relationship into a natural descriptive clause (e.g., “the university that she graduated from,” “the city where that university is located”).  
4. **Conclude Clearly:** End with a specific question that asks for the `{answer_name}` type.  
5. **Final Review:** Ensure the question sounds natural, logical, and conceals all hidden entities.

---

## EXAMPLES (For Your Reference, please study them carefully before generating the question)

Example 1:
Fact Path(s):
Kyloq Espinay(Person) -> current_living_city -> Belltower(City)
Start Node: Kyloq Espinay
Answer Name: Belltower

Excellent Question:
Kyloq Espinay is updating a travel profile and needs the current residence listed; in which city is Kyloq Espinay currently living?

Bad Questions:
Where does Kyloq Espinay live—Belltower, right?  (Reveals the answer)
What is Kyloq Espinay's location?  (Too generic; not explicitly a city)
Which city is Belltower?  (Asks about the answer itself rather than eliciting it)

Example 2:
Fact Path(s):
Kestrel (City) -> number_of_universities -> 1 (University Count)
Start Node: Kestrel
Answer Name: 1

Excellent Question:
Kestrel is preparing an education overview; state the number of universities located in Kestrel.

Bad Questions:
Which city has 1 university?  (Does not begin with the start node and could match multiple cities)
How many universities are in Kestrel—1, correct?  (Leads with or implies the answer)
What is the figure associated with Kestrel?  (Too vague; not clearly about universities)

Example 3:
Fact Path(s):
114801502 (Franchise Revenue) -> franchise_revenue_of -> Riptide Propulsion (Company)
Start Node: 114801502
Answer Name: Riptide Propulsion

Excellent Question:
A business analyst was reviewing revenue reports and found a figure of 114,801,502 dollars attributed to a particular franchise. Which company generated that amount in franchise revenue?

Bad Question Examples:

“What is the franchise revenue of Riptide Propulsion?”  (Reveals the answer; wrong direction.)
“Which franchise made 114801502?”  (Unnatural phrasing, lacks context.)
“Riptide Propulsion earned how much?”  (Inverts the path; asks the wrong thing.)

You will now receive the inputs. Generate the question. Make sure to follow all the CRITICAL RULES above.
\end{VerbatimWrap}
\end{tcolorbox}

%% file: prompts/simple_qa_scenario_2.tex
\begin{tcolorbox}[colback=lightbluebg!30!white,colframe=blueframe,breakable,title=Prompt for Simple QA Scenario 2 hop]
\begin{VerbatimWrap}
You are a master question designer specializing in crafting complex, {path_type} reasoning questions to benchmark advanced AI systems. Your goal is to transform a structured "fact path" into a single, natural, high-quality question whose answer is the final node in that path.

## CONTEXT & INPUTS

You will receive:
- **Path Type:** The number of "hops" (1-hop, 2-hop, 3-hop, etc.).
- **Fact Path(s):** A structured list of linked facts in the form (Entity A) --[relationship]--> (Entity B).
- **Answer Name:** The final entity in the path — the correct answer to your question.

Inputs provided:
Fact Path(s):
{facts_string}
Intermediate Nodes: {intermediate_nodes}
Start Node: {start_node}
Answer Name: {answer_name}

---

## CRITICAL RULES

1. **Strictly Grounded:** Use *only* the provided Fact Path(s). Do not invent new entities, names, events, or relationships.  
2. **Conceal Hidden Nodes:** Never reveal the `{answer_name}` or any `{intermediate_nodes}` directly in the question.  
3. **Logical Sequence:** The question must begin with the **Start Node** and follow the exact order of relationships step by step.  
4. **Natural Scenario:** Wrap the relationships into a short, natural scenario or story. It should feel realistic and engaging, but not overly elaborate.  
5. **Specific Questioning:** Avoid vague question words like “what” or “where.” Instead, specify the category of the answer (e.g., “In which city…”, “At which company…”).  
6. **Explicit Ask:** The question must clearly request the final answer entity (the `{answer_name}`), without ambiguity.  
7. **Final Output:** Output *only* the question text — no preambles, labels, or explanations.

---

## QUESTION DESIGN PROCESS

1. **Deconstruct the Path:** Identify the start, intermediates, and final nodes, and the relationships connecting them.  
2. **Anchor the Start Node:** Begin the question with the `{start_node}`.  
3. **Weave the Narrative:** Convert each relationship into a natural descriptive clause (e.g., “the university that she graduated from,” “the city where that university is located”).  
4. **Conclude Clearly:** End with a specific question that asks for the `{answer_name}` type.  
5. **Final Review:** Ensure the question sounds natural, logical, and conceals all hidden entities.

---

## EXAMPLES (For Your Reference, please study them carefully before generating the question)

Example 1: 2-hop
Fact Path(s):
(37031562782510(GDP)) --[gdp_of]--> (Tulvir(Country))
(Tulvir(Country)) --[number_of_ethnic_groups]--> (10(Ethnic Group Count))
Intermediate Nodes:
[Tulvir]
Start Node:
37031562782510
Answer Name:
10

Excellent Question:
The GDP figure of 37031562782510 is being reviewed by a development economist compiling a diversity brief; after determining the country associated with this GDP figure, what is the number of ethnic groups recorded for that country?

Bad Question:
1. Tulvir has a GDP of 37031562782510—how many ethnic groups are there? (Reveals the intermediate node)
2. 37031562782510 is a GDP figure; name the country. (Stops at the first hop; omits the second relationship)
3. In which country are there 10 ethnic groups? (Leads with the answer value and ignores the start node and first hop)
4. Provide the number of ethnic groups in Tulvir. (Reveals intermediate node and skips the first hop logic from the start node)

Example 2: 2-hop
Fact Path(s):
(Phyla Krasnic(Person)) --[current_working_company]--> (Bioscape Terraforming(Company))
(Bioscape Terraforming(Company)) --[number_of_departments]--> (43(Department Count))
Intermediate Nodes:
[Bioscape Terraforming]
Start Node:
Phyla Krasnic
Answer Name:
43

Excellent Question:
Phyla Krasnic is preparing an internal orientation packet; for the company where Phyla Krasnic is currently employed, please find the total number of departments.

Bad Question:
1. What is the number of departments in the company where Phyla Krasnic is currently employed? (Not a natural question with scenario)
2. How many departments does Bioscape Terraforming have? (Reveals the intermediate node)
3. Phyla Krasnic works somewhere; how many departments are there? (Missing the second-hop anchor that the departments belong to the employer identified via the first hop; too vague)

Example 3: 2-hop
Fact Path(s):
(Shattrath(City)) --[sister_city_of]--> (Volantis(City))
(Volantis(City)) --[located_country]--> (Ikthos(Country))
Intermediate Nodes:
[Volantis]
Start Node:
Shattrath
Answer Name:
Ikthos

Excellent Question:
Shattrath is updating its cultural partnerships page; for the city that holds a sister-city relationship with Shattrath, name the country in which that partner city is located.

Bad Question:
1. Shattrath has a sister city; which city is it? (Stops after the first hop and does not reach the required answer type)
2. Name the sister city of Shattrath that is in Ikthos. (Reveals the answer and asks for the wrong entity)
3. Shattrath has cultural ties; name the country of its main partner. (Vague; skips the explicit sister-city relationship step)

You will now receive the inputs. Generate the question. Make sure to follow all the CRITICAL RULES above.
\end{VerbatimWrap}
\end{tcolorbox}

%% file: prompts/simple_qa_scenario_3.tex
\begin{tcolorbox}[colback=lightbluebg!30!white,colframe=blueframe,breakable,title=Prompt for Simple QA Scenario 3 hop]
\begin{VerbatimWrap}
You are a master question designer specializing in crafting complex, {path_type} reasoning questions to benchmark advanced AI systems. Your goal is to transform a structured "fact path" into a single, natural, high-quality question whose answer is the final node in that path.

## CONTEXT & INPUTS

You will receive:
- **Path Type:** The number of "hops" (1-hop, 2-hop, 3-hop, etc.).
- **Fact Path(s):** A structured list of linked facts in the form (Entity A) --[relationship]--> (Entity B).
- **Answer Name:** The final entity in the path — the correct answer to your question.

Inputs provided:
Fact Path(s):
{facts_string}
Intermediate Nodes: {intermediate_nodes}
Start Node: {start_node}
Answer Name: {answer_name}

---

## CRITICAL RULES

1. **Strictly Grounded:** Use *only* the provided Fact Path(s). Do not invent new entities, names, events, or relationships.  
2. **Conceal Hidden Nodes:** Never reveal the `{answer_name}` or any `{intermediate_nodes}` directly in the question.  
3. **Logical Sequence:** The question must begin with the **Start Node** and follow the exact order of relationships step by step.  
4. **Natural Scenario:** Wrap the relationships into a short, natural scenario or story. It should feel realistic and engaging, but not overly elaborate.  
5. **Specific Questioning:** Avoid vague question words like “what” or “where.” Instead, specify the category of the answer (e.g., “In which city…”, “At which company…”).  
6. **Explicit Ask:** The question must clearly request the final answer entity (the `{answer_name}`), without ambiguity.  
7. **Final Output:** Output *only* the question text — no preambles, labels, or explanations.

---

## QUESTION DESIGN PROCESS

1. **Deconstruct the Path:** Identify the start, intermediates, and final nodes, and the relationships connecting them.  
2. **Anchor the Start Node:** Begin the question with the `{start_node}`.  
3. **Weave the Narrative:** Convert each relationship into a natural descriptive clause (e.g., “the university that she graduated from,” “the city where that university is located”).  
4. **Conclude Clearly:** End with a specific question that asks for the `{answer_name}` type.  
5. **Final Review:** Ensure the question sounds natural, logical, and conceals all hidden entities.

---

## EXAMPLES (For Your Reference, please study them carefully before generating the question)

Example 1: 3-hop
Fact Path(s):
(Brondar Drakaar(Person)) --[spouse]--> (Kryll Dornwald(Person))
(Kryll Dornwald(Person)) --[current_living_city]--> (Thunder(City))
(Thunder(City)) --[number_of_universities]--> (1(University Count))
Intermediate Nodes:
[Kryll Dornwald, Thunder]
Start Node:
Brondar Drakaar
Answer Name:
1

Excellent Question:
Brondar Drakaar’s biographer is tracing how family ties relate to local education access; beginning with Brondar Drakaar’s spouse and the city where that spouse currently lives, please find the total number of universities in that city.

Bad Question:
1. How many universities are in Thunder? (Reveals an intermediate node; doesn’t begin with the start node)
2. Starting from Brondar’s city, count universities. (Skips the spouse step; incorrect path order)
3. Name the spouse of Brondar Drakaar and their city. (Asks for intermediates, not the final answer)
4. Brondar Drakaar’s spouse lives in Thunder; confirm there is 1 university. (Reveals the intermediate city and presupposes the answer)

Example 2: 3-hop
Fact Path(s):
(14730(Employee Count)) --[number_of_employees_of]--> (Starfall Exoplanetary(Company))
(Starfall Exoplanetary(Company)) --[headquarter_city]--> (Evermore(City))
(Evermore(City)) --[located_city_of]--> (Oakhart Seminary(University))
Intermediate Nodes:
[Starfall Exoplanetary, Evermore]
Start Node:
14730
Answer Name:
Oakhart Seminary

Excellent Question:
A corporate analyst starts from a headcount figure of 14730 to identify the firm it belongs to; after determining that company's headquarters city, what is the university located in that city.

Bad Question:
1. Which university is in Evermore? (Reveals an intermediate node; ignores the required chain from the start node)
2. From the HQ city, find the company with 14730 staff. (Reverses the path and asks for extra information)
3. For the university in that city, how many employees are there? (Asks for the wrong target and muddles roles)

Example 3: 3-hop
Fact Path(s):
(Karthos(Country)) --[capital_city]--> (Cinderhollow(City))
(Cinderhollow(City)) --[sister_city]--> (Cerulea(City))
(Cerulea(City)) --[mayor_of_city_for]--> (Zennith Folara(Person))
Intermediate Nodes:
[Cinderhollow, Cerulea]
Start Node:
Karthos
Answer Name:
Zennith Folara

Excellent Question:
A person is preparing an international relations briefing; starting from Karthos's capital, then identifying that capital's sister city, what is the name of the person who serves as mayor of that sister city.

Bad Question:
1. Who is the mayor of Cerulea? (Reveals an intermediate node; skips the ordered chain from the start node)
2. Karthos has a capital; name it. (Stops after the first hop; does not reach the final entity)
3. Name the capital and its sister city. (Asks for intermediates; not the final answer)

You will now receive the inputs. Generate the question. Make sure to follow all the CRITICAL RULES above.
\end{VerbatimWrap}
\end{tcolorbox}

%% file: prompts/parallel_same_numerical_sum.tex
\begin{tcolorbox}[colback=lightbluebg!30!white,colframe=blueframe,breakable,title=Prompt for Parallel QA Same Numerical Sum]
\begin{VerbatimWrap}
You are a master question designer specializing in creating a parallel question based on two sub-questions whose answers are the same type of numerical information.

## CONTEXT

You will be given two sub-questions and their answers. Your goal is to combine the two sub-questions into a single parallel question, where the answers of the sub-questions are of the same type and are numerical, e.g. Year, GDP, Population Count, and you should make up a new question that asks the summation of the answers of the two sub-questions in a natural way.

## INPUT FORMAT

You will receive the following inputs:
*   **Question 1:** [The first question] | **Answer:** [The answer to Q1]
*   **Question 2:** [The second question] | **Answer:** [The answer to Q2]

Your Inputs:
Sub Questions:
Question 1: {sub_question1} Answer: {sub_answer1} ({sub_answer_type1})
Question 2: {sub_question2} Answer: {sub_answer2} ({sub_answer_type2})

## CRITICAL RULES

1.  **Analyze the Questions:** You will receive two questions (Question 1, Question 2) and their answers.
2.  **Remove Scenario Descriptions:** IMPORTANT - When combining the sub-questions, you MUST remove any scenario descriptions, context explanations, or situational details from the original questions. Only keep the core question structure and entity information.
3.  **Identify the Link:** Based on the answers and questions, combine the two sub-questions and make up a new question that asks the summation of the answers of the two sub-questions in a natural way.
4.  **Keep the orginal logic:** While connecting the two sub-questions, you should strictly keep the original logic of the each sub-questions unchanged, but without scenario descriptions.
5.  **Final Output Format:** Your final output must be a JSON object with the following format:
    {
        "Question": "The combined question",
        "Answer": "The answer to the combined question"
    }

## Example

### Your Input:

*   **Question 1:** Khepran Quintoris is filling out a biographical form and needs to specify his birth year; in which year was Khepran Quintoris born? | **Answer:** 1968 (Year)
*   **Question 2:** Xantheus Varden is updating his biography and needs to include his birth year; in which year was Xantheus Varden born? | **Answer:** 1978 (Year)

### Your Required Output:
```json
{{
    "Question": "What is the summation of the birth years of Khepran Quintoris and Xantheus Varden?",
    "Answer": "3946"
}}
```

## Additional Example

### Your Input:

*   **Question 1:** What is the GDP of the country whose capital is City A? | **Answer:** 10000 (GDP)
*   **Question 2:** How much is the GDP of the country whose leader is the person whose spouse is person X? | **Answer:** 20000 (GDP)

### Your Required Output:
```json
{{
    "Question": "What is the summation of the GDP of the country whose leader is the person whose spouse is person X and the GDP of the country whose capital is City A?",
    "Answer": "30000"
}}
```
\end{VerbatimWrap}
\end{tcolorbox}

%% file: prompts/parallel_same_numerical_difference.tex
\begin{tcolorbox}[colback=lightbluebg!30!white,colframe=blueframe,breakable,title=Prompt for Parallel QA Same Numerical Difference]
\begin{VerbatimWrap}
You are a master question designer specializing in creating a parallel question based on two sub-questions whose answers are the same type of entity.

## CONTEXT

You will be given two sub-questions and their answers that are of the same type of entity, together with the numerical attribute of the answer entities. Your goal is to combine the two sub-questions into a single parallel question, where the answers of the sub-questions are of the same type of entity, e.g. Person, Company, City, University, and you should make up a new question that asks the ABSOLUTE DIFFERENCE of the numerical attributes of the answer entities in a natural way.

## INPUT FORMAT

You will receive the following inputs:
*   **Question 1:** [The first question] | **Answer:** [The answer to Q1] | **Answer Entity Type:** [The type of the answer entity] | **Answer Entity Attribute Relation:** [The attribute relation of the answer entity] | **Attribute Value:** [The numerical value of the attribute]
*   **Question 2:** [The second question] | **Answer:** [The answer to Q2] | **Answer Entity Type:** [The type of the answer entity] | **Answer Entity Attribute Relation:** [The attribute relation of the answer entity] | **Attribute Value:** [The numerical value of the attribute]

Your Inputs:
Sub Questions:
Question 1: {sub_question1} | Answer: {sub_answer1} | Answer Entity Type: {sub_answer_entity_type1} | Answer Entity Attribute Relation: {sub_answer_entity_attribute_relation1} | Attribute Value: {sub_answer_entity_attribute_value1}
Question 2: {sub_question2} | Answer: {sub_answer2} | Answer Entity Type: {sub_answer_entity_type2} | Answer Entity Attribute Relation: {sub_answer_entity_attribute_relation2} | Attribute Value: {sub_answer_entity_attribute_value2}

## CRITICAL RULES

1.  **Analyze the Questions:** You will receive two questions (Question 1, Question 2), their answers and numerical attributes of the answer entities.
2.  **Remove Scenario Descriptions:** IMPORTANT - When combining the sub-questions, you MUST remove any scenario descriptions, context explanations, or situational details from the original questions. Only keep the core question structure and entity information.
3.  **Identify the Link:** Based on the answers, numerical attributes of answer entities and questions, combine the two sub-questions and make up a new question that asks the ABSOLUTE DIFFERENCE of the numerical attributes of the answer entities in a natural way.
4.  **Keep the original logic:** While connecting the two sub-questions, you should strictly keep the original logic of each sub-question unchanged, but without scenario descriptions.
5.  **Final Output Format:** Your final output must be a JSON object with the following format:
    {
        "Question": "The combined question",
        "Answer": "The answer to the combined question"
    }

## Examples

### Example 1: Difference Operation

### Your Input:

*   Question 1: "Sisyphon Kimaro is updating a biographical form and needs to specify his best friend's information; who is the best friend of Sisyphon Kimaro?" | Answer: Morphax Serpentine | Answer Entity Type: Person | Answer Entity Attribute Relation: Morphax Serpentine -> birth_year -> 1987 | Attribute Value: 1987
*   Question 2: "Ruolan Ravenscar is preparing a social network map and needs to identify connections; who is the best friend of Ruolan Ravenscar?" | Answer: Lycomed Septimus | Answer Entity Type: Person | Answer Entity Attribute Relation: Lycomed Septimus -> birth_year -> 1964 | Attribute Value: 1964

### Your Required Output:
```json
{{
    "Question": "What is the absolute difference in birth years between the best friend of Sisyphon Kimaro and the best friend of Ruolan Ravenscar?",
    "Answer": "23"
}}
```

### Example 2: Difference Operation with Complex Questions

### Your Input:

*   Question 1: "Starting from the country Sorgan, identify the best friend of the mayor of its sister country's capital city's sister city." | Answer: Person A | Answer Entity Type: Person | Answer Entity Attribute Relation: Person A -> birth_year -> 1980 | Attribute Value: 1980
*   Question 2: "Which person serves as the mayor of the sister city of the city in which the university Kronus Takamura attended is located?" | Answer: Person B | Answer Entity Type: Person | Answer Entity Attribute Relation: Person B -> birth_year -> 1964 | Attribute Value: 1964

### Your Required Output:
```json
{{
    "Question": "What is the absolute difference in birth years between the best friend of the mayor of country Sorgan's sister country's capital city's sister city and the mayor of the sister city of the city where Kronus Takamura attended university?",
    "Answer": "16"
}}
```
\end{VerbatimWrap}
\end{tcolorbox}

%% file: prompts/parallel_same_entity_compare.tex
\begin{tcolorbox}[colback=lightbluebg!30!white,colframe=blueframe,breakable,title=Prompt for Parallel Same Entity Compare]
\begin{VerbatimWrap}
You are a master question designer specializing in creating a parallel question based on two sub-questions whose answers are the same type of entity.

## CONTEXT

You will be given two sub-questions and their answers that are of the same type of entity, together with the numerical attribute of the answer entities. Your goal is to combine the two sub-questions into a single parallel question, where the answers of the sub-questions are of the same type of entity, e.g. Person, Company, City, University, and you should make up a new question that compares the attributes of the answer entities in a natural way.

## INPUT FORMAT

You will receive the following inputs:
*   **Question 1:** [The first question] | **Answer:** [The answer to Q1] | **Answer Entity Type:** [The type of the answer entity] | **Answer Entity Attribute:** [The attribute of the answer entity]
*   **Question 2:** [The second question] | **Answer:** [The answer to Q2] | **Answer Entity Type:** [The type of the answer entity] | **Answer Entity Attribute:** [The attribute of the answer entity]

Your Inputs:
Sub Questions:
Question 1: {sub_question1} | Answer: {sub_answer1} | Answer Entity Type: {sub_answer_entity_type1} | Answer Entity Attribute Relation: {sub_answer_entity_attribute_relation1} | Attribute Value: {sub_answer_entity_attribute_value1}
Question 2: {sub_question2} | Answer: {sub_answer2} | Answer Entity Type: {sub_answer_entity_type2} | Answer Entity Attribute Relation: {sub_answer_entity_attribute_relation2} | Attribute Value: {sub_answer_entity_attribute_value2}

## CRITICAL RULES

1.  **Analyze the Questions:** You will receive two questions (Question 1, Question 2), their answers and attributes of the answer entities.
2.  **Identify the Link:** Based on the answers, attributes of answer entities and questions, combine the two sub-questions and make up a new question that asks the comparison of the attributes of the answer entities in a natural way.
3.  **Keep the orginal logic:** While connecting the two sub-questions, you should strictly keep the original logic of the each sub-questions unchanged.
4.  **Final Output Format:** Your final output must be a JSON object with the following format:
    {
        "Question": "The combined question",
        "Answer": "The answer to the combined question"
    }

## Example

### Your Input:

*   Question 1: "What is the mayor of the city where the Person A was born?" | Answer: Person B | Answer Entity Type: Person | Answer Entity Attribute Relation: Person B -> birth_year -> 1968 | Attribute Value: 1968
*   Question 1: "Who is the leader of the sister country to the country where the headquarter city of Company X is located?" | Answer: Person C | Answer Entity Type: Person | Answer Entity Attribute Relation: Person C -> birth_year -> 1966 | Attribute Value: 1966

### Your Required Output:
```json
{{
    "Question": "Who is older, the mayor of the city where the Person A was born, or the leader of the sister country to the country where the headquarter city of Company X is located?",
    "Answer": "Person C"
}}
```
\end{VerbatimWrap}
\end{tcolorbox}

%% file: prompts/parallel_same_entity_sum.tex
\begin{tcolorbox}[colback=lightbluebg!30!white,colframe=blueframe,breakable,title=Prompt for Parallel Same Entity Sum]
\begin{VerbatimWrap}
You are a master question designer specializing in creating a parallel question based on two sub-questions whose answers are the same type of entity.

## CONTEXT

You will be given two sub-questions and their answers that are of the same type of entity, together with the numerical attribute of the answer entities. Your goal is to combine the two sub-questions into a single parallel question, where the answers of the sub-questions are of the same type of entity, e.g. Person, Company, City, University, and you should make up a new question that asks the SUM of the numerical attributes of the answer entities in a natural way.

## INPUT FORMAT

You will receive the following inputs:
*   **Question 1:** [The first question] | **Answer:** [The answer to Q1] | **Answer Entity Type:** [The type of the answer entity] | **Answer Entity Attribute Relation:** [The attribute relation of the answer entity] | **Attribute Value:** [The numerical value of the attribute]
*   **Question 2:** [The second question] | **Answer:** [The answer to Q2] | **Answer Entity Type:** [The type of the answer entity] | **Answer Entity Attribute Relation:** [The attribute relation of the answer entity] | **Attribute Value:** [The numerical value of the attribute]

Your Inputs:
Sub Questions:
Question 1: {sub_question1} | Answer: {sub_answer1} | Answer Entity Type: {sub_answer_entity_type1} | Answer Entity Attribute Relation: {sub_answer_entity_attribute_relation1} | Attribute Value: {sub_answer_entity_attribute_value1}
Question 2: {sub_question2} | Answer: {sub_answer2} | Answer Entity Type: {sub_answer_entity_type2} | Answer Entity Attribute Relation: {sub_answer_entity_attribute_relation2} | Attribute Value: {sub_answer_entity_attribute_value2}

## CRITICAL RULES

1.  **Analyze the Questions:** You will receive two questions (Question 1, Question 2), their answers and numerical attributes of the answer entities.
2.  **Remove Scenario Descriptions:** IMPORTANT - When combining the sub-questions, you MUST remove any scenario descriptions, context explanations, or situational details from the original questions. Only keep the core question structure and entity information.
3.  **Identify the Link:** Based on the answers, numerical attributes of answer entities and questions, combine the two sub-questions and make up a new question that asks the SUM of the numerical attributes of the answer entities in a natural way.
4.  **Keep the original logic:** While connecting the two sub-questions, you should strictly keep the original logic of each sub-question unchanged, but without scenario descriptions.
5.  **Final Output Format:** Your final output must be a JSON object with the following format:
    {
        "Question": "The combined question",
        "Answer": "The answer to the combined question"
    }

## Examples

### Example 1: Sum Operation

### Your Input:

*   Question 1: "Sisyphon Kimaro is updating a biographical form and needs to specify his best friend's information; who is the best friend of Sisyphon Kimaro?" | Answer: Morphax Serpentine | Answer Entity Type: Person | Answer Entity Attribute Relation: Morphax Serpentine -> birth_year -> 1987 | Attribute Value: 1987
*   Question 2: "Ruolan Ravenscar is preparing a social network map and needs to identify connections; who is the best friend of Ruolan Ravenscar?" | Answer: Lycomed Septimus | Answer Entity Type: Person | Answer Entity Attribute Relation: Lycomed Septimus -> birth_year -> 1964 | Attribute Value: 1964

### Your Required Output:
```json
{{
    "Question": "What is the summation of the birth years of the best friend of Sisyphon Kimaro and the best friend of Ruolan Ravenscar?",
    "Answer": "3951"
}}
```

### Example 2: Difference Operation with Complex Questions

### Your Input:

*   Question 1: "Starting from the country Sorgan, identify the best friend of the mayor of its sister country's capital city's sister city." | Answer: Person A | Answer Entity Type: Person | Answer Entity Attribute Relation: Person A -> birth_year -> 1980 | Attribute Value: 1980
*   Question 2: "Which person serves as the mayor of the sister city of the city in which the university Kronus Takamura attended is located?" | Answer: Person B | Answer Entity Type: Person | Answer Entity Attribute Relation: Person B -> birth_year -> 1964 | Attribute Value: 1964

### Your Required Output:
```json
{{
    "Question": "What is the absolute difference in birth years between the best friend of the mayor of country Sorgan's sister country's capital city's sister city and the mayor of the sister city of the city where Kronus Takamura attended university?",
    "Answer": "16"
}}
```
\end{VerbatimWrap}
\end{tcolorbox}

%% file: prompts/parallel_same_entity_difference.tex
\begin{tcolorbox}[colback=lightbluebg!30!white,colframe=blueframe,breakable,title=Prompt for Parallel Same Entity Sum]
\begin{VerbatimWrap}
You are a master question designer specializing in creating a parallel question based on two sub-questions whose answers are the same type of entity.

## CONTEXT

You will be given two sub-questions and their answers that are of the same type of entity, together with the numerical attribute of the answer entities. Your goal is to combine the two sub-questions into a single parallel question, where the answers of the sub-questions are of the same type of entity, e.g. Person, Company, City, University, and you should make up a new question that asks the ABSOLUTE DIFFERENCE of the numerical attributes of the answer entities in a natural way.

## INPUT FORMAT

You will receive the following inputs:
*   **Question 1:** [The first question] | **Answer:** [The answer to Q1] | **Answer Entity Type:** [The type of the answer entity] | **Answer Entity Attribute Relation:** [The attribute relation of the answer entity] | **Attribute Value:** [The numerical value of the attribute]
*   **Question 2:** [The second question] | **Answer:** [The answer to Q2] | **Answer Entity Type:** [The type of the answer entity] | **Answer Entity Attribute Relation:** [The attribute relation of the answer entity] | **Attribute Value:** [The numerical value of the attribute]

Your Inputs:
Sub Questions:
Question 1: {sub_question1} | Answer: {sub_answer1} | Answer Entity Type: {sub_answer_entity_type1} | Answer Entity Attribute Relation: {sub_answer_entity_attribute_relation1} | Attribute Value: {sub_answer_entity_attribute_value1}
Question 2: {sub_question2} | Answer: {sub_answer2} | Answer Entity Type: {sub_answer_entity_type2} | Answer Entity Attribute Relation: {sub_answer_entity_attribute_relation2} | Attribute Value: {sub_answer_entity_attribute_value2}

## CRITICAL RULES

1.  **Analyze the Questions:** You will receive two questions (Question 1, Question 2), their answers and numerical attributes of the answer entities.
2.  **Remove Scenario Descriptions:** IMPORTANT - When combining the sub-questions, you MUST remove any scenario descriptions, context explanations, or situational details from the original questions. Only keep the core question structure and entity information.
3.  **Identify the Link:** Based on the answers, numerical attributes of answer entities and questions, combine the two sub-questions and make up a new question that asks the ABSOLUTE DIFFERENCE of the numerical attributes of the answer entities in a natural way.
4.  **Keep the original logic:** While connecting the two sub-questions, you should strictly keep the original logic of each sub-question unchanged, but without scenario descriptions.
5.  **Final Output Format:** Your final output must be a JSON object with the following format:
    {
        "Question": "The combined question",
        "Answer": "The answer to the combined question"
    }

## Examples

### Example 1: Difference Operation

### Your Input:

*   Question 1: "Sisyphon Kimaro is updating a biographical form and needs to specify his best friend's information; who is the best friend of Sisyphon Kimaro?" | Answer: Morphax Serpentine | Answer Entity Type: Person | Answer Entity Attribute Relation: Morphax Serpentine -> birth_year -> 1987 | Attribute Value: 1987
*   Question 2: "Ruolan Ravenscar is preparing a social network map and needs to identify connections; who is the best friend of Ruolan Ravenscar?" | Answer: Lycomed Septimus | Answer Entity Type: Person | Answer Entity Attribute Relation: Lycomed Septimus -> birth_year -> 1964 | Attribute Value: 1964

### Your Required Output:
```json
{{
    "Question": "What is the absolute difference in birth years between the best friend of Sisyphon Kimaro and the best friend of Ruolan Ravenscar?",
    "Answer": "23"
}}
```

### Example 2: Difference Operation with Complex Questions

### Your Input:

*   Question 1: "Starting from the country Sorgan, identify the best friend of the mayor of its sister country's capital city's sister city." | Answer: Person A | Answer Entity Type: Person | Answer Entity Attribute Relation: Person A -> birth_year -> 1980 | Attribute Value: 1980
*   Question 2: "Which person serves as the mayor of the sister city of the city in which the university Kronus Takamura attended is located?" | Answer: Person B | Answer Entity Type: Person | Answer Entity Attribute Relation: Person B -> birth_year -> 1964 | Attribute Value: 1964

### Your Required Output:
```json
{{
    "Question": "What is the absolute difference in birth years between the best friend of the mayor of country Sorgan's sister country's capital city's sister city and the mayor of the sister city of the city where Kronus Takamura attended university?",
    "Answer": "16"
}}
```
\end{VerbatimWrap}
\end{tcolorbox}

%% file: prompts/combo_qa.tex
\begin{tcolorbox}[colback=lightbluebg!30!white,colframe=blueframe,breakable,title=Prompt for Combo QA]
\begin{VerbatimWrap}
You are a master question designer. Your task is to rewrite a second question (Question 2) to make it entirely dependent on the answer of a first question (Question 1).

## CONTEXT

You will be given two sequential questions. The answer to Question 1 is a piece of information required to answer Question 2. Your goal is to rephrase Question 2 by replacing the starting information with a dependent reference to the answer of Question 1.

## INPUT FORMAT

You will receive the following inputs:
*   **Question 1:** [The first question] | **Answer:** [The answer to Q1]
*   **Question 2:** [The second question, which contains the answer to Q1] | **Answer:** [The answer to Q2]

Your Inputs:
Sub Questions: 
Question 1: {sub_question1} Answer: {sub_answer1} ({sub_answer_type1})
Question 2: {sub_question2} Answer: {sub_answer2} ({sub_answer_type2})

## YOUR TASK & CRITICAL RULES

1.  **Analyze the Questions:** You will receive two questions (Question 1, Question 2) and their answers.
2.  **Identify the Link:** Find the specific piece of information (the "linking entity") in Question 2 that is identical to the answer of Question 1.
3.  **Rephrase Question 2:** Rewrite Question 2 by replacing the "linking entity" with a generic, dependent reference. This reference MUST point back to the answer of Question 1.
    *   **Examples of dependent references:** "the person from the answer of the first question," "the city you obtained from the answer of the first question"
4.  **Preserve Intent:** The rephrased question must follow the exact same logic as the original Question 2. Do not alter its core meaning and logic, your job is only to replace the starting information with a dependent reference to the answer of Question 1.

## EXAMPLE

### Example 1:

### Your Input:

*   **Question 1:** How many faculties does the university located in the birth city of Jackson Barton's spouse have? | **Answer:** 3 (Faculty Count)
*   **Question 2:** Who is the mother of the leader of the country where the city that headquarters a company with 3 products is located? | **Answer:** Devereux Bryan (Person)

### Your Required Output:
Example 1: Who is the mother of the leader of the country where the city that headquarters a company with n products is located, where n is the **numerical answer** to the first question.
Example 2: Who is the mother of the leader of the country where the city that headquarters a company with a number of products **equal to the answer of the first question** is located?

### Example 2:

### Your Input:

*   **Question 1:** What is the birth year of the mother of the person who is the mayor of the city where the university that has 179 research patents is located? | **Answer:** 1980 (Year)
*   **Question 2:** In which city is the company founded in the same year you obtained in Question 1 headquartered? | **Answer:** Zarnok (City)

### Your Required Output:
In which city is the company founded in the same year that you obtained from Question 1 headquartered?

## FINAL OUTPUT INSTRUCTIONS

Your final output must be the raw text of the slightly rephrased Question 2.**
*   Do NOT include any preamble, titles, or labels (like "Rephrased Question:").
*   Do NOT include explanations or markdown.
\end{VerbatimWrap}
\end{tcolorbox}

%% file: prompts/llm_as_judge.tex
\begin{tcolorbox}[colback=lightbluebg!30!white,colframe=blueframe,breakable,title=LLM-as-a-Judge Prompt]
\begin{VerbatimWrap}
You are an evaluation assistant. Please determine if the predicted answer is equivalent to the labeled answer.

Question: {question}

Labeled Answer: {gt_answer}

Predicted Answer: {pred_answer}

Did the model give an answer **equivalent** to the labeled answer? Please respond with "Correct" if they are equivalent, or "Incorrect" if they are not equivalent.

The output should in the following json format:
```json
{{
    "rationale": your rationale for the judgement, as a text,
    "judgement": your judgement result, can only be "Correct" or "Incorrect",
}}
```
\end{VerbatimWrap}
\end{tcolorbox}

%% file: prompts/llm_data_quality.tex
\begin{tcolorbox}[colback=lightbluebg!30!white,colframe=blueframe,breakable,title=LLM Data Quality Assessment Prompt]
\begin{VerbatimWrap}
You are a question quality evaluator. Please evaluate the quality of the given question based on the following criteria:

Question: {question}
Ground Truth Answer: {golden_answer}

Please evaluate the question on these 4 criteria (each should be scored as 0-1, where 1 means the issue exists):

1. **Factual Error**: Does the question contain factual errors or incorrect premises?
   - Score 1 if the question has factual errors
   - Score 0 if the question is factually correct
   - Example: "What is Dennis Rodman's occupation?" with GT "Actor" - The question assumes Dennis Rodman is known for being an actor, which is factually incorrect (he's primarily a basketball player)

2. **Time Sensitive**: Is the question time-sensitive (does the answer change over time)?
   - Score 1 if the question is time-sensitive
   - Score 0 if the question is stable over time
   - Example: "Who is the current CEO of Apple?" - Answer changes over time
   - Example: "Who was the producer of Today?" - Unclear which "Today" (show, song, etc.) and when

3. **Mix Language**: Does the question contain mixed languages?
   - Score 1 if the question mixes multiple languages
   - Score 0 if the question uses a single language
   - Example: Question contains non-English text mixed with English, like Russian words

4. **Clarity**: Is the question itself unclear or hard to understand?
   - Score 1 if the question is unclear, ambiguous, or difficult to understand
   - Score 0 if the question is clear and well-formed
   - Consider grammar, wording, and overall comprehensibility

Please provide your evaluation in the following JSON format:
```json
{{
    "factual_error": <score 0 or 1>,
    "time_sensitive": <score 0 or 1>,
    "mix_language": <score 0 or 1>,
    "clarity": <score 0 or 1>,
    "rationale": {{
        "factual_error": "<explanation>",
        "time_sensitive": "<explanation>",
        "mix_language": "<explanation>",
        "clarity": "<explanation>"
    }}
}}
```

IMPORTANT: Only output the JSON, nothing else.
\end{VerbatimWrap}
\end{tcolorbox}